\documentclass[10pt,twocolumn,letterpaper]{article}

\usepackage[pagenumbers]{wacv} 
\usepackage{graphicx}

\usepackage[dvipsnames,table,xcdraw]{xcolor}
\usepackage{tikz}
\usepackage{comment}
\usepackage{amsmath,amssymb} 
\usepackage{color}
\usepackage{array}
\usepackage{float}
\usepackage{tabularx}
\usepackage{multicol}
\usepackage{multirow}
\usepackage{booktabs}
\usepackage{nicematrix}
\usepackage{colortbl}
\usepackage{amssymb}
\usepackage{pifont}%
\newcommand{\xmark}{\text{\ding{55}}}
\usepackage{tikz}
\usepackage{makecell}

\usepackage[pagebackref,breaklinks,colorlinks]{hyperref}
\usepackage{adjustbox}

\usepackage[capitalize]{cleveref}
\crefname{section}{Sec.}{Secs.}
\Crefname{section}{Section}{Sections}
\Crefname{table}{Table}{Tables}
\crefname{table}{Tab.}{Tabs.}

\def\wacvPaperID{2360} 
\def\confName{WACV}
\def\confYear{2024}
\def\NAME{\texttt{CLIP-DIY}\xspace}

\usepackage[normalem]{ulem}

\usepackage{dsfont}

\newif\ifshowedits

\newcommand{\addeditor}[3]{%
  \definecolor{#1color}{rgb}{#3}
  \expandafter\newcommand\csname #1\endcsname[1]{
  \ifshowedits
    {\color{#1color}##1}
  \else
    {##1}
  \fi
  }%
  \expandafter\newcommand\csname #1c\endcsname[1]{
  \ifshowedits
    {\color{#1color}{\bf[#2: ##1]}}
  \else
    {}
  \fi
  }%
}

\newcommand{\newtvar}[1]{
  \expandafter\newcommand\csname #1\endcsname{\text{#1}}
}

\newlength{\fHeight}
\newcommand{\faBicycle}{\includegraphics[height=\fHeight]{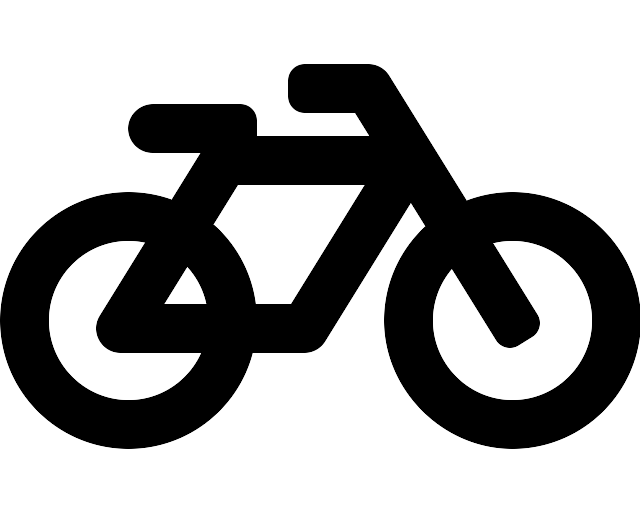}}
\newcommand{\faBird}{\includegraphics[height=\fHeight]{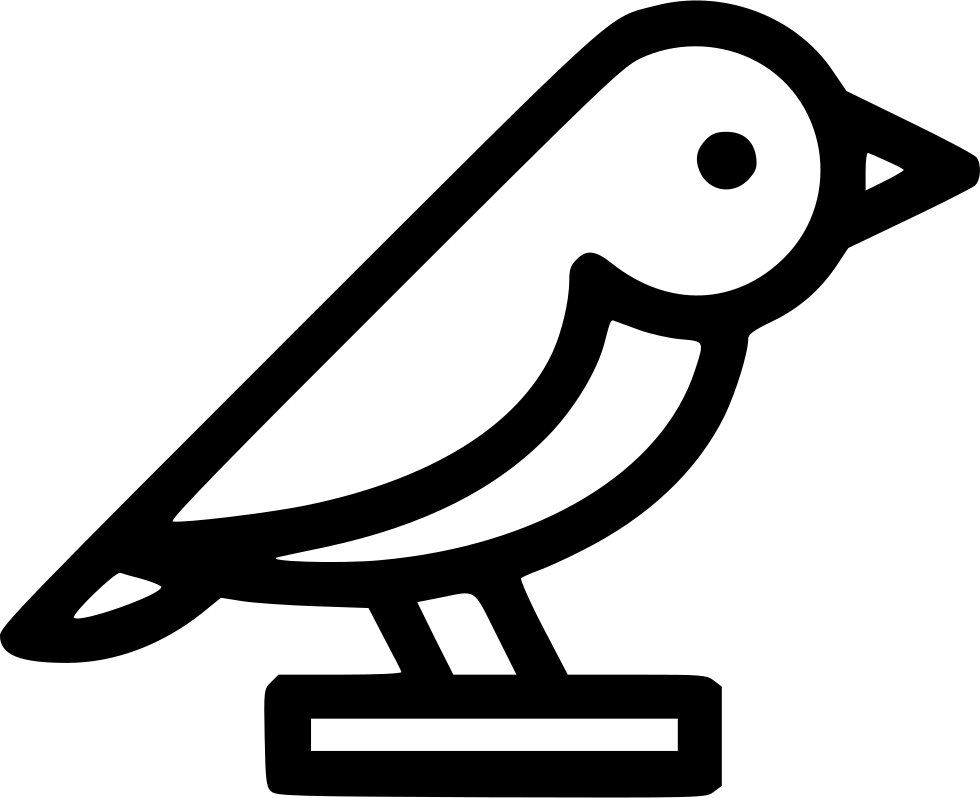}}
\newcommand{\faBus}{\includegraphics[height=\fHeight]{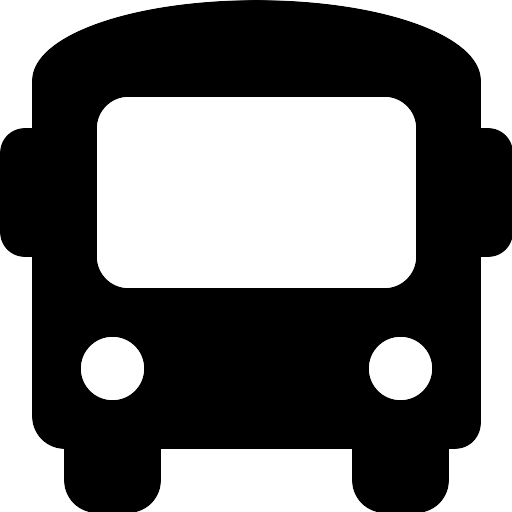}}
\newcommand{\faCar}{\includegraphics[height=\fHeight]{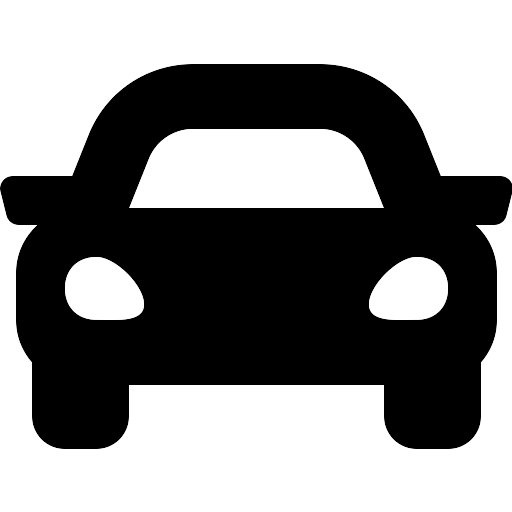}}
\newcommand{\faCat}{\includegraphics[height=\fHeight]{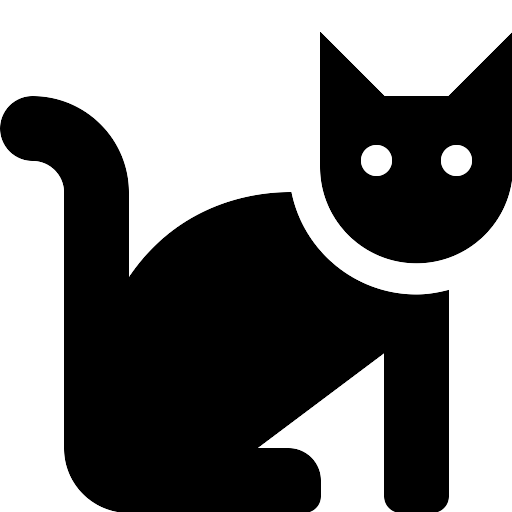}}
\newcommand{\faCow}{\includegraphics[height=\fHeight]{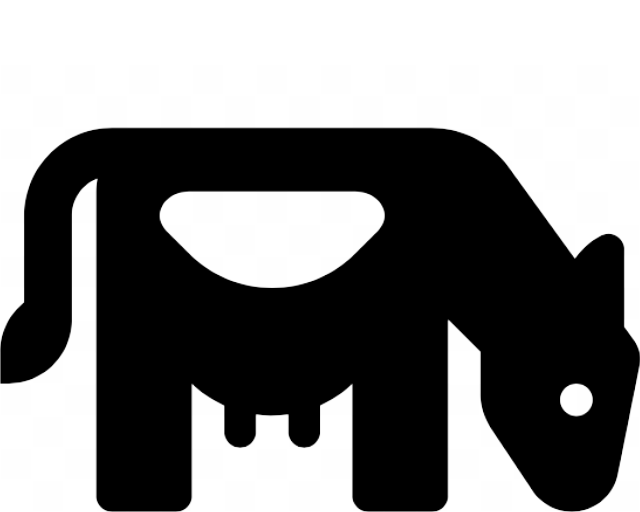}}
\newcommand{\faChair}{\includegraphics[height=\fHeight]{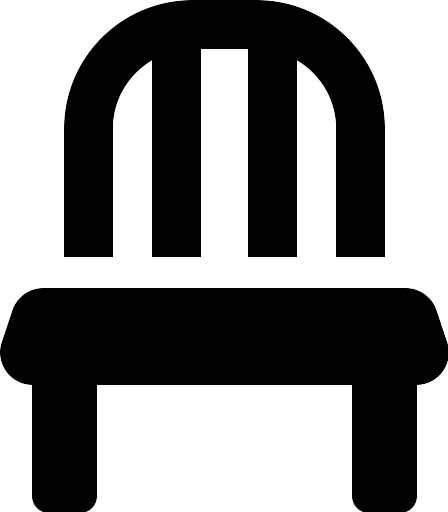}}
\newcommand{\faCouch}{\includegraphics[height=\fHeight]{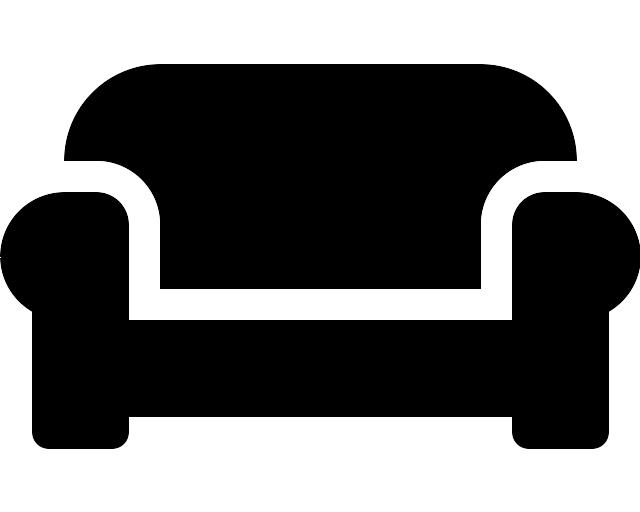}}

\newcommand{\faDog}{\includegraphics[height=\fHeight]{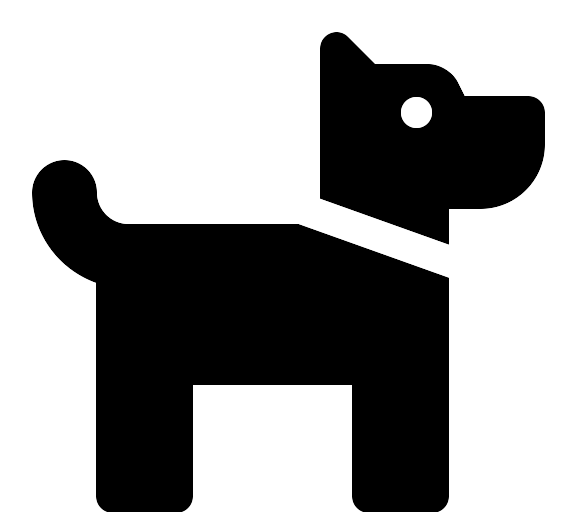}}
\newcommand{\faDinningTable}{\includegraphics[height=\fHeight]{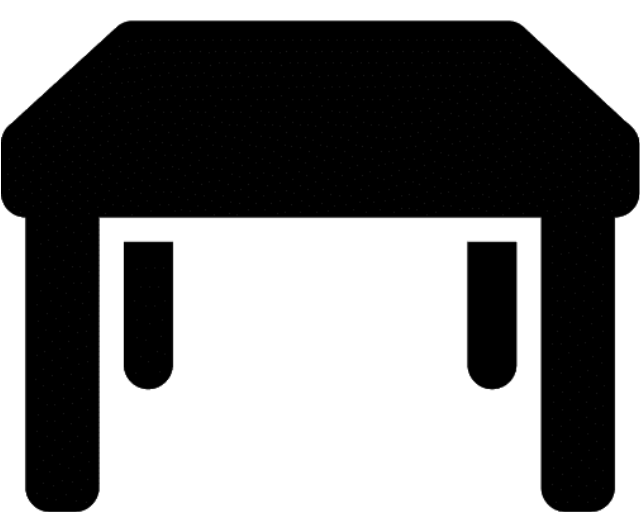}}
\newcommand{\faHorse}{\includegraphics[height=\fHeight]{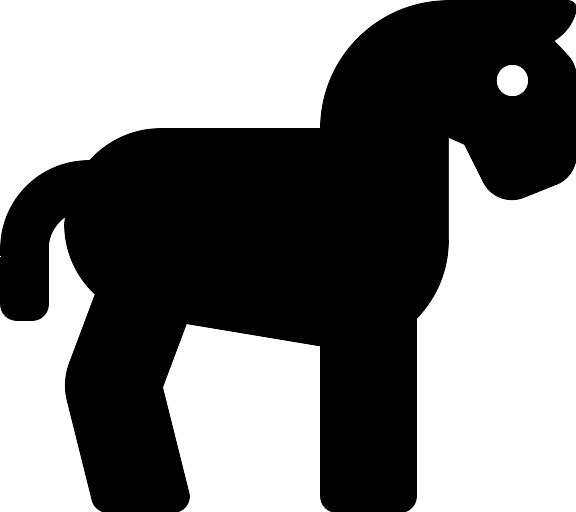}}
\newcommand{\faMotorcycle}{\includegraphics[height=\fHeight]{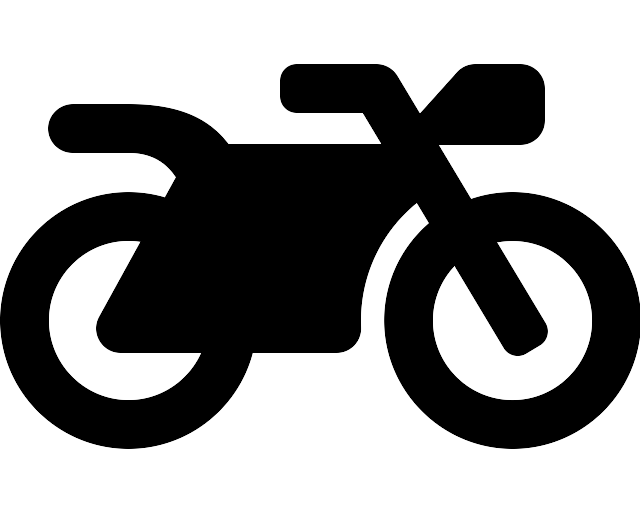}}
\newcommand{\faPlane}{\includegraphics[height=\fHeight]{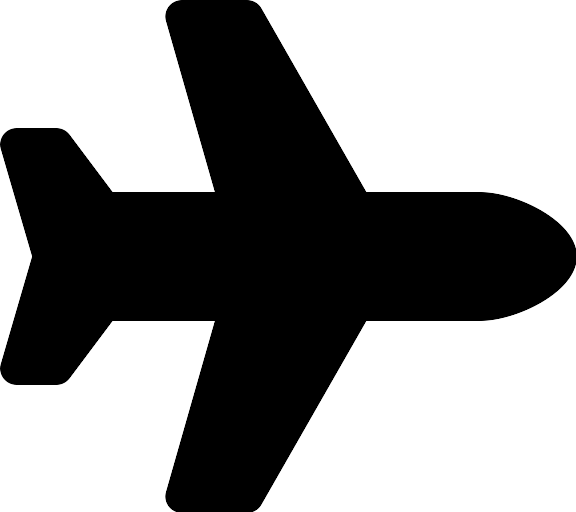}}
\newcommand{\faShip}{\includegraphics[height=\fHeight]{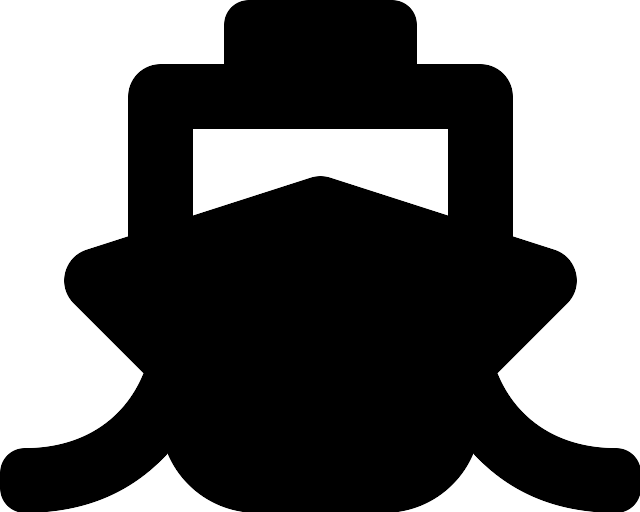}}
\newcommand{\faSheep}{\includegraphics[height=\fHeight]{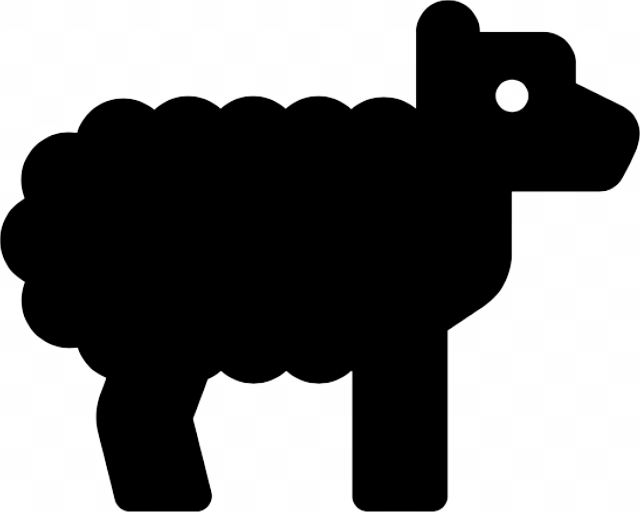}}
\newcommand{\faBackground}{\includegraphics[height=\fHeight]{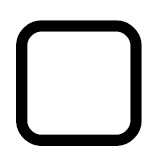}}
\newcommand{\faTrain}{\includegraphics[height=\fHeight]{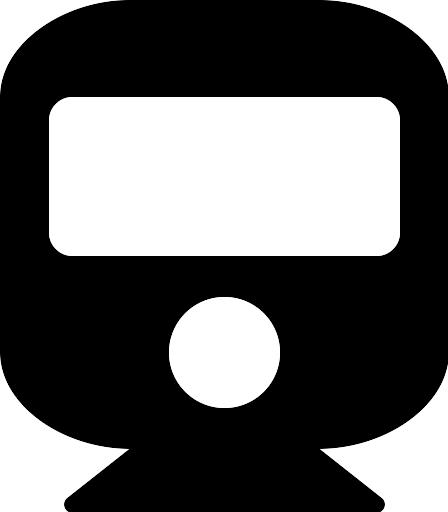}}
\newcommand{\faTulip}{\includegraphics[height=\fHeight]{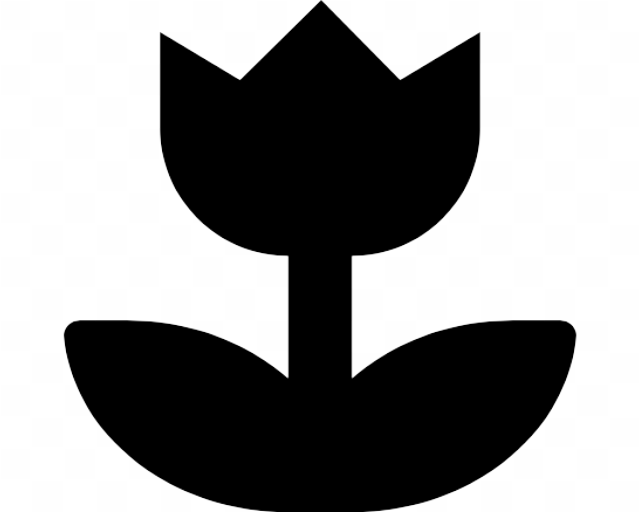}}
\newcommand{\faTv}{\includegraphics[height=\fHeight]{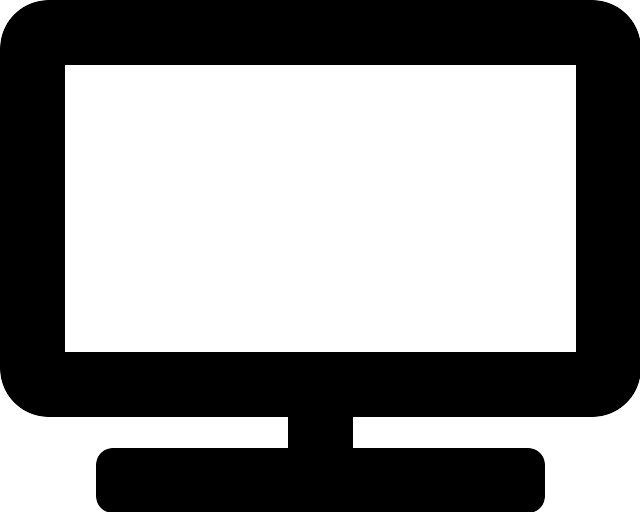}}
\newcommand{\faWalking}{\includegraphics[height=\fHeight]{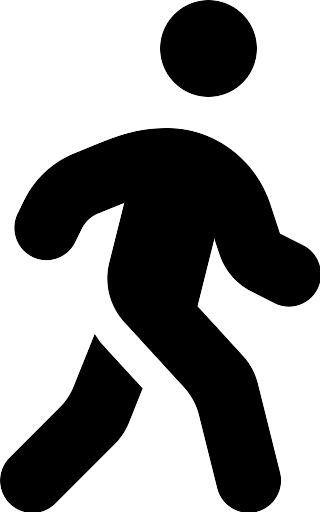}}
\newcommand{\faWineBottle}{\includegraphics[height=\fHeight]{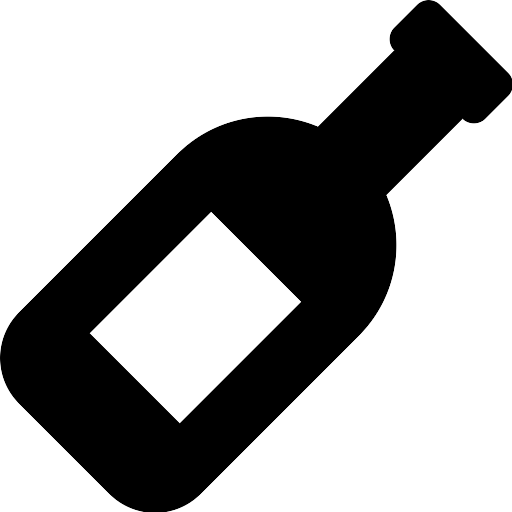}}
\newcommand{\tableIcons}{%
\faBackground & \faPlane & \faBicycle & \faBird & \faShip & \faWineBottle & \faBus & \faCar & \faCat &	\faChair & \faCow & \faDinningTable & \faDog & \faHorse & \faMotorcycle & \faWalking & \faTulip & \faSheep & \faCouch & \faTrain & \faTv}
\definecolor{cow}{RGB}{64, 128, 0}
\definecolor{sheep}{RGB}{128, 64, 0}
\definecolor{train}{RGB}{128, 192, 0}
\definecolor{dog}{RGB}{64, 0, 128}
\definecolor{sofa}{RGB}{0, 192, 0}
\definecolor{surf}{RGB}{0, 224, 128}
\definecolor{umbrella}{RGB}{0, 160, 0}
\definecolor{elephant}{RGB}{128, 128, 0}
\definecolor{bear}{RGB}{200, 0, 32}
\definecolor{bird_coco}{RGB}{128, 0, 110}
\definecolor{zebra}{RGB}{128, 32, 0}
\definecolor{remote}{RGB}{128, 192, 192}
\definecolor{fork}{RGB}{64, 0, 160}
\definecolor{bowl}{RGB}{192, 0, 32}
\definecolor{cellphone}{RGB}{192, 160, 128}
\definecolor{sink}{RGB}{128, 64, 128}
\definecolor{hair}{RGB}{64, 128, 224}
\definecolor{bird_pascal}{RGB}{128, 128, 0}
\definecolor{aeroplane}{RGB}{190, 0, 0}
\definecolor{bottle}{RGB}{128, 0, 128}
\definecolor{tv/monitor}{RGB}{0, 64, 128}
\definecolor{bottle}{RGB}{200, 0, 200}
\definecolor{chair}{RGB}{192, 0, 0}
\definecolor{plant}{RGB}{192, 32, 128}
\definecolor{bench}{RGB}{0, 160, 192}
\definecolor{orange}{RGB}{64, 224, 0}
\definecolor{vase}{RGB}{192, 128, 224}
\definecolor{scissors}{RGB}{255, 140, 31}
\definecolor{thread}{RGB}{10, 206, 250}
\definecolor{apple}{RGB}{0, 128, 128}
\definecolor{bench_coco}{RGB}{0, 200, 200}
\definecolor{darkgreen}{RGB}{0, 200, 20}
\definecolor{dumplings}{RGB}{128, 0, 128}
\definecolor{bus}{RGB}{0, 128, 128}
\definecolor{boat}{RGB}{0, 0, 128}
\definecolor{plant_pascal}{RGB}{0, 64, 0}
\definecolor{bike}{RGB}{0, 128, 0}
\definecolor{horse}{RGB}{128, 128, 224}
\definecolor{cow_coco}{RGB}{128, 160, 192}
\definecolor{pizza}{RGB}{0, 140, 220}
\definecolor{spoon}{RGB}{128, 128, 64}
\definecolor{cake}{RGB}{192, 224, 0}
\definecolor{bowl}{RGB}{192, 0, 32}
\definecolor{microwave}{RGB}{128, 192, 0}
\definecolor{refrigerator}{RGB}{64, 128, 96}
\definecolor{sheep_coco}{RGB}{0, 0, 192}
\definecolor{sandwich}{RGB}{0, 128, 32}
\definecolor{cup}{RGB}{128, 128, 128}
\definecolor{knife}{RGB}{128, 224, 128}
\definecolor{chair_coco}{RGB}{0, 128, 64}
\definecolor{dog_coco}{RGB}{0, 32, 192}
\definecolor{cat}{RGB}{128, 0, 192}
\definecolor{carrot}{RGB}{80, 80, 150}
\definecolor{diningtable}{RGB}{192, 128, 0}
\definecolor{nemo}{RGB}{255, 80, 31}
\definecolor{dory}{RGB}{10, 170, 250}
\definecolor{greyeleph}{RGB}{100, 170, 150}
\definecolor{pinkeleph}{RGB}{243, 10, 198}
\definecolor{pasteis}{RGB}{0, 70, 150}
\definecolor{coffee}{RGB}{143, 107, 8}
\definecolor{scrabble}{RGB}{10, 206, 250}
\definecolor{pencil}{RGB}{255, 140, 31}
\definecolor{rubber stamp}{RGB}{55, 240, 31}
\definecolor{paper clip}{RGB}{200, 0, 25}

\newcommand{\clipimg}{\Phi_{V}}
\newcommand{\cliptxt}{\Phi_{T}}
\newcommand{\clipmap}{\rho}
\newcommand{\scales}{\mathcal{S}}
\newcommand{\queries}{\mathcal{T}}
\newcommand{\saliency}{\hat{M}}
\newcommand{\clipmask}{M_{CLIP}}
\newcommand{\partition}{\mathcal{X}}
\newcommand{\promptstyle}[2]{\textbf{\texttt{\color{#1}#2}}}
\DeclareMathOperator*{\upsample}{\text{Upsample}}
\DeclareMathOperator*{\softmax}{\text{SoftMax}}

\begin{document}
\title{\texttt{CLIP-}\texttt{\textcolor{dumplings}{DIY}}: \texttt{CLIP} \textcolor{dumplings}{\texttt{D}}ense \textcolor{dumplings}{\texttt{I}}nference \textcolor{dumplings}{\texttt{Y}}ields Open-Vocabulary Semantic Segmentation For-Free}

\author{
Monika Wysoczańska$^1$\thanks{Corresponding author: 
    \texttt{monika.wysoczanska.dokt@pw.edu.pl}}
\and
Michaël Ramamonjisoa$^2$
\and
Tomasz Trzciński$^{1,4,5}$
\and 
Oriane Siméoni$^3$
\\
  $^1$Warsaw University of Technology, $^2$Meta AI, $^3$Valeo.ai,
  $^4$Tooploox, $^5$IDEAS NCBR
  }
\maketitle

\begin{abstract}

The emergence of CLIP has opened the way for \emph{open-world} image perception. The zero-shot classification capabilities of the model are impressive but are harder to use for dense tasks such as image segmentation. 
Several methods have proposed different modifications and learning schemes to produce dense output. 
Instead, we propose in this work an open-vocabulary semantic segmentation method, dubbed \NAME, 
which does not require any additional training or annotations, but instead leverages existing unsupervised object localization approaches.
In particular, \NAME is a multi-scale approach that directly exploits CLIP classification abilities on patches of different sizes and aggregates the decision in a single map. We further guide the segmentation using foreground/background scores obtained using unsupervised object localization methods. With our method, we obtain state-of-the-art zero-shot semantic segmentation results on PASCAL VOC and perform on par with the best methods on COCO.

\end{abstract}

\newlength{\teaserheight}
\setlength{\teaserheight}{2.5cm}
\begin{figure}
\centering
\begin{tabular}{
cc
}
     \begin{subfigure}[m]{0.44\linewidth}
         \centering
         \includegraphics[width=\textwidth, height=\teaserheight]{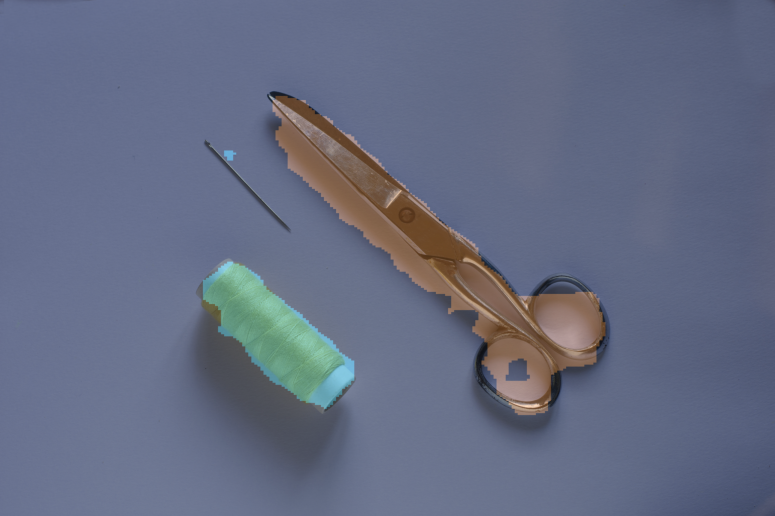}
     \end{subfigure}
     & 
     \begin{subfigure}[m]{0.44\linewidth}
         \centering
         \includegraphics[width=\textwidth, height=\teaserheight]{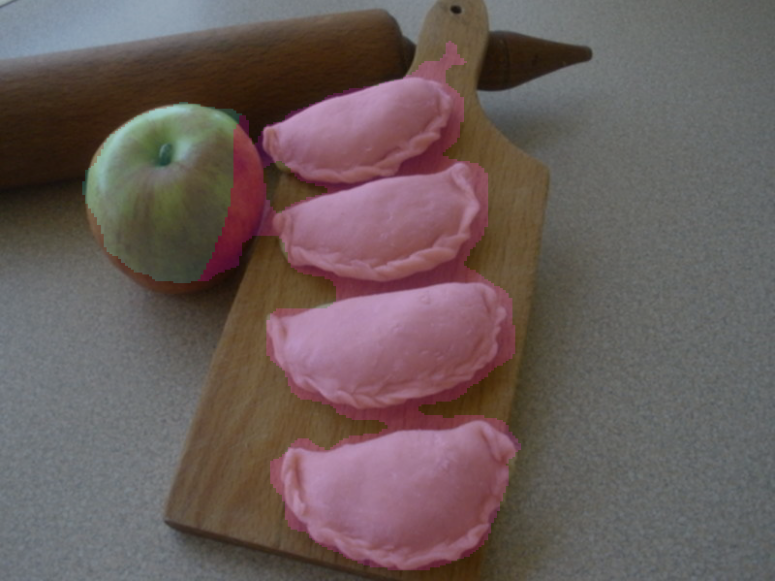}
     \end{subfigure}\hspace{-1.1em}
     \\
     \shortstack[lm]{\promptstyle{scissors}{metal scissors} \\ \promptstyle{thread}{thread} } &
     \shortstack[lm]{\color{dumplings}\textbf{\texttt{polish dumplings}} \\ \text{\color{apple}\textbf{\texttt{apple}}} } \\ 
     
     \multicolumn{2}{l}{
     \begin{subfigure}[m]{0.98\linewidth}
         \centering
         \includegraphics[width=\textwidth, height=3cm,trim=10 60 10 30, clip]{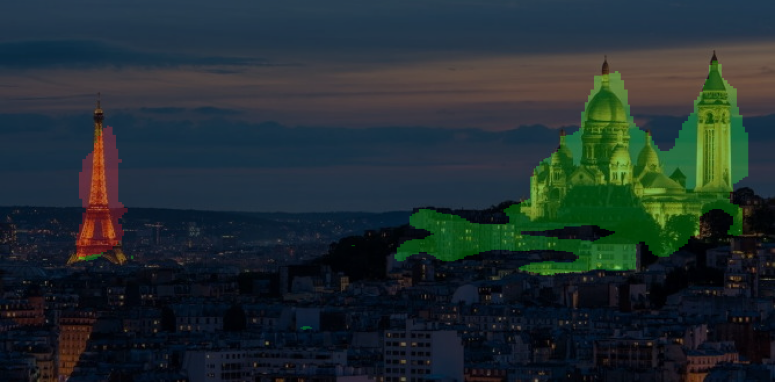}
     \end{subfigure}} \\
     \multicolumn{2}{c}{
     \centering
     \promptstyle{red}{Eiffel tower},
     \promptstyle{darkgreen}{Sacré coeur}
     }
\end{tabular}
    \vspace{-1em}
    \caption{
    \textbf{Simple and accurate semantic segmentation for-free using} \NAME. Our method processes regions of an image \emph{separately}, \emph{in parallel} to produce per-patch-per-class scores, given a set of prompts that can take \emph{any length}. \NAME does not require re-training, and can therefore immediately adapt to any new vocabulary.
    }
    \vspace{-1em}
    \label{fig:teaser}
\end{figure}

\section{Introduction}
\label{sec:intro}

The task of \emph{semantic segmentation}, which aims at predicting the class of every pixel in an image has been widely tackled using fully-supervised approaches~\cite{hao2020brief, strudel2021segmenter,NEURIPS2021_950a4152}, 
which require tedious and therefore expensive per-pixel annotations.
Moreover, semantic segmentation has typically been performed with a \emph{finite set of classes}~\cite{pascal-voc-2012, behley2019semantickitti, lin2014microsoft} describing the types of objects that should be discovered in images. However, using a fixed number of classes is limiting for real-world applications as interesting object classes may vary in time and per application -- having to perform annotation and re-training on new classes is expensive and sub-optimal. In this context, recent advances in Visual Language Models (VLMs)~\cite{openclip, alayrac2022flamingo, jia2021scaling} have paved the way to \emph{open-vocabulary} perception. 
Indeed, VLMs, trained with cheap and widely available image-text pairs, e.g. captions, offer new possibilities to describe images with a \emph{large and open vocabulary}. Using such models can help alleviate both the problem of supervision and finite vocabulary. 

In particular, the popular 
VLM Open-CLIP~\cite{openclip} has been exploited to perform \emph{open-vocabulary perception}. While it has high performance on image classification, applying OpenCLIP on dense tasks is more challenging~\cite{zhou2022maskclip}. In order to improve CLIP segmentation abilities, different methods have been proposed to modify the architecture \cite{zhou2022maskclip, clippy2022}, to add new modules \cite{mukhoti2023pacl, liu2022open, cha2022tcl} or to train new specifically designed models \cite{clippy2022,xu2022groupvit} from scratch. 
Instead, we propose \NAME, a new \emph{zero-shot open-vocabulary semantic segmentation} approach which makes direct use of the high-performance image classification properties of CLIP, \emph{does not} need architecture changes or additional training. In particular, our method applies CLIP to a multi-scale grid of patches and aggregates the information into a single prediction map.

Moreover, in order to further improve the quality of the localization of our predicted maps, we propose with \NAME to leverage the recent efforts in \emph{unsupervised object localization}. This task aims at discovering every object depicted in images in a class-agnostic fashion and thus \emph{without manual annotation}. Recent methods \cite{simeoni2021lost, wang2022tokencut, wang2023cut, shin2022selfmask, bielski2022move, simeoni2023found} achieve impressive localization results by leveraging self-supervised features \cite{caron2021dino, mocov2}. While some methods discover one object per image~\cite{simeoni2021lost,wang2022tokencut}, more recent ones try to highlight all objects in an image~\cite{shin2022selfmask, wang2023cut, simeoni2023found}. 

We therefore propose to leverage those capabilities in \NAME, by guiding CLIP predictions with a very lightweight unsupervised foreground/background strategy which greatly improves the predictions' quality. 

To summarize, our novel approach \NAME best leverages 
the open-world classification capabilities of CLIP and the high-quality of unsupervised object localization approaches yielding the following contributions:

\begin{itemize}
    \item We introduce \NAME, a novel, simple technique for open-vocabulary semantic segmentation which \emph{does not require additional training} or any \emph{pixel-level annotation} but instead leverages strong self-supervised features with good localization properties combined with CLIP.
    \item Our multi-scale approach which uses simply CLIP
    as it was designed ---for image classification--- enables \NAME to produce well-localized predictions. 
    \item We demonstrate that unsupervised foreground/background methods can be effectively used to provide spatial guidance to CLIP predictions. 
    \item We achieve a new state-of-the-art zero-shot open-vocabulary semantic segmentation on PASCAL VOC dataset and perform on par with the best methods on COCO. 
    \item We perform an extensive validation of the design of our method, and show that it is robust as it can be directly applied to in-the-wild open-world segmentation.

\end{itemize}


\section{Related work}
In this section, we discuss previous work related to ours, starting in Sec.~\ref{ssec:related_zeroshot} with zero-shot open-vocabulary semantic methods, then following with unsupervised object localization methods in Sec.~\ref{ssec:unsupervised_obj}. Finally, in Section~\ref{ssec:ssl_clip} we focus more specifically on works that leverage a combination of self-supervised learned features and CLIP to perform open-world segmentation.

\subsection{Zero-shot open-vocabulary semantic segmentation}
\label{ssec:related_zeroshot}

With the aim to build generalizable models, \emph{zero-shot} methods for semantic segmentation \cite{xian2019semantic,zhao2017open,bucher2019neurips,gu2020acmmm,xian2019semantic,kato2019zero,hu2020uncertainty,li2020consistent,pastore2021closer} 
propose to extend models trained in a fully supervised fashion on a set of \emph{seen} classes to new \emph{unseen} classes. Many leverage
relationships encoded in pre-trained word embeddings \cite{word2vec, GloVe} to discover new unseen concepts. Such methods require annotation for the seen classes while we aim to use no pixel-level annotation.

Alternatively, \emph{open-vocabulary} approaches \cite{wu2023open} exploit image-text alignment without needing to pre-define vocabulary. Several \cite{zhou2022maskclip,lou2022segclip,mukhoti2023pacl,rao2021denseclip, lou2022segclip} build on top of the popularCLIP~\cite{radford2021learning, openclip} model which showed impressive \emph{global} text-image alignment properties but lacks localization quality~\cite{negclip}. 
Using class-agnostic object masks, it is possible to learn to align the embeddings of selected pixels with text~\cite{rao2021denseclip, ghiasi2022openseg, ding2021ZegFormer}, but at the cost of pixel-level annotations. 
Without extra supervision, MaskCLIP \cite{zhou2022maskclip} alters the last pooling layer of CLIP 
to produce dense predictions and use them as pseudo-labels to train a segmentation model, forming MaskCLIP+.
Using only image captions---cheap to acquire and widely available---as supervision, \cite{xu2022groupvit,lou2022segclip,liu2022open,xu2023learning} learn local alignment between image \emph{regions} and paired text with contrastive objectives.
 Regions are formed using a learnt hierarchical mechanism~\cite{xu2022groupvit}, cross-attention based clustering~\cite{lou2022segclip}, using a clustering head trained with diverse view~\cite{liu2022open} or slot-attention~\cite{xu2023learning}.
PACL~\cite{mukhoti2023pacl} adds an embedder module that learns affinity between patches and the global text token, TCL~\cite{cha2022tcl} builds a new local contrastive objective which directly aligns captions with pre-selected patches and ViewCO\cite{ren2023viewco} proposes a multi-view consistent learning approach. CLIPpy \cite{clippy2022} proposes to fully re-train CLIP with a few well-designed modifications to directly obtained denser features.
Alternatively, ReCO~\cite{shin2022reco} builds prototypes of the desired vocabulary (using CLIP-based retrieval) which are then used for co-segmentation. 

Rather than modifying the architecture of CLIP or training a new module specifically designed to densify its outputs, we propose to directly use \emph{as is} the good classification ability of the model. Indeed, we perform a dense multi-scale patch classification. By doing so, our method can easily be adapted to any new dataset, with any new vocabulary.

\subsection{Unsupervised object localization}
\label{ssec:unsupervised_obj}
\begin{figure*}[h]
    \centering    \includegraphics[ width=\textwidth]{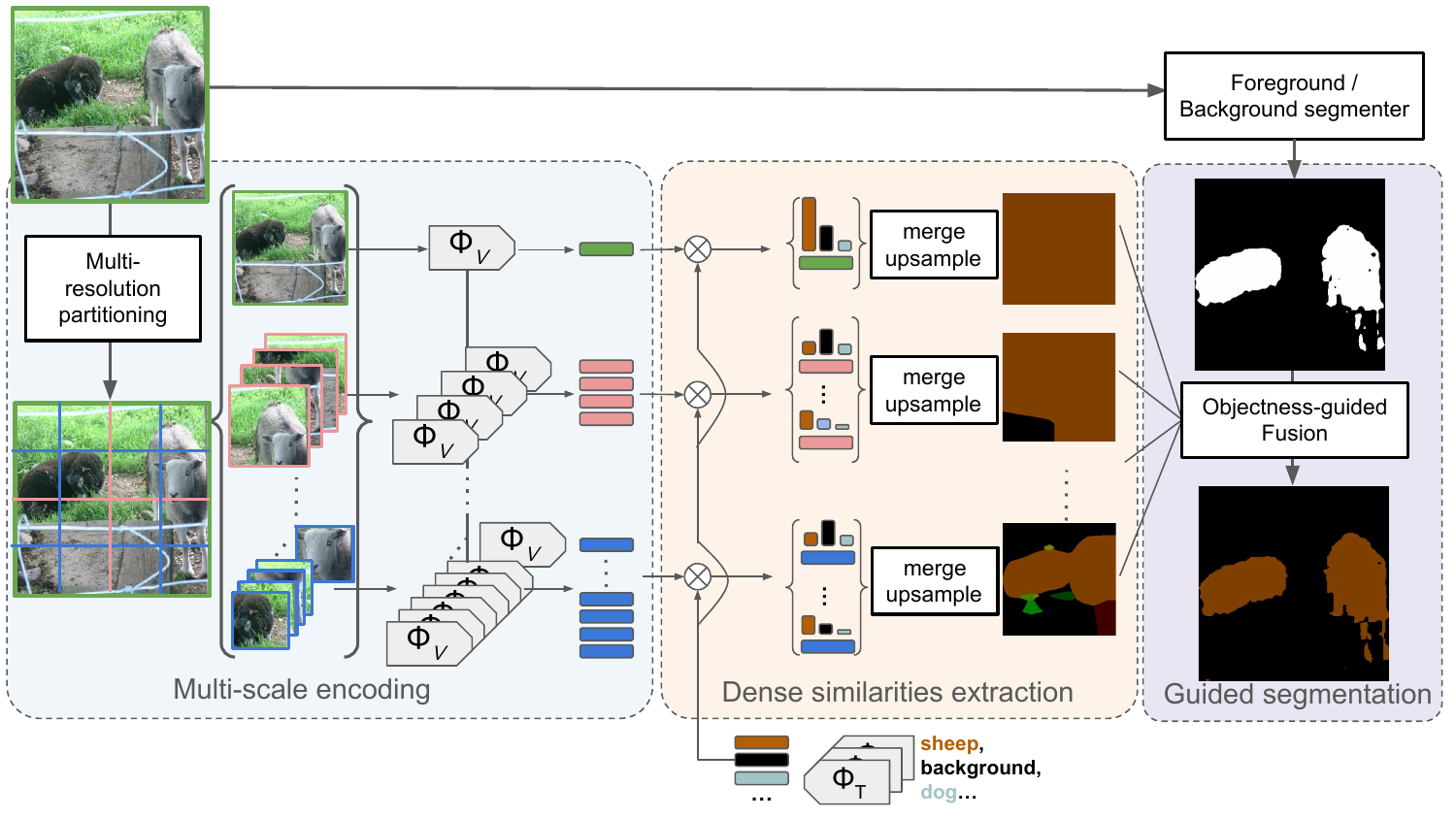}
    \vspace{-2em}
    \caption{\textbf{Overview of \NAME}, our two-step pipeline for segmentation map extraction. We note $\clipimg$ and $\cliptxt$ the CLIP encoders for image and text respectively.
    (1) An input image is partitioned into smaller patches, and each of them is fed \emph{indepently} to the image encoder,  yielding a vector of per-class similarity to an \emph{arbitrary-length} vocabulary of classes. (2) Patches are then aggregated back before upsampling to produce dense similarity maps. (3) An objectness score obtained by an off-the-shelf foreground-background segmentation method such as FOUND~\cite{simeoni2023found} is used to guide the prediction of the final segmentation.}
    \label{fig:overview}
\end{figure*} 

Interestingly, recent works have shown that ViT features trained in a self-supervised fashion \cite{caron2020swav,caron2021dino,mocov2} on images--with no human-made annotations--have good localization properties \cite{simeoni2021lost, wang2022tokencut}. Such properties have been exploited to tackle the problem of \emph{unsupervised object localization} \cite{Vo20rOSD,vo2021largescale} which requires to localize objects--any object--depicted in images, and such without any cue. A set of methods~\cite{simeoni2021lost,wang2022tokencut,melas2022deepsectralmethod, shin2022selfmask, wang2023cut} exploit the good correlation properties of the feature and find an object as the set of patches which highly differs to the other patches. Alternatively~\cite{wang2022freesolo}, exploits the attention mechanisms with different queries and produces maps that are ranked and filtered and~\cite{simeoni2023found} proposes to look for the background instead of the objects in order to avoid single object discovery and to need priors about objects. The coarse object localization results obtained using those methods can be used as pseudo-labels to train large instance or segmentation models in a class-agnostic fashion~\cite{simeoni2021lost, wang2022tokencut,wang2022freesolo, zadaianchuk2022comus,wang2023cut}. Recent FOUND~\cite{simeoni2023found} is a very light model--a single conv1x1--trained to produce foreground/background segmentation of good quality. When self-trained FOUND achieves even better results and discovers more objects per image~\cite{simeoni2021lost,wang2023cut}. In this work, we propose to leverage the good object localization properties of unsupervised object localization models, which make no hypotheses about object classes and remain therefore \emph{open}. In particular, we take advantage of the efficiency of FOUND~\cite{simeoni2023found} to guide our zero-shot segmentation.

\subsection{Combining self-supervised features \& CLIP}
\label{ssec:ssl_clip}

Combining self-supervised learning~\cite{caron2021dino,he2020moco, mocov2,he2022masked} with VLMs has been previously explored by different open-vocabulary segmentation methods \cite{shin2022reco, clippy2022, Rewatbowornwong2023ZeroGuideSeg}. Correlation qualities are used to perform co-segmentation~\cite{shin2022reco} when pre-training properties are directly leveraged to initialize the visual encoder backbone~\cite{clippy2022,xu2023learning}. 
Related to our work, ZGS~\cite{Rewatbowornwong2023ZeroGuideSeg} builds potential masks using clustering strategies on self-supervised features and assigns them a class. Contrary to all previous approaches, it explores the task without predefined text prompts. Although ZGS currently obtains lower results than other baselines on the task, it opens an interesting direction for future work.

In this work, we do not perform co-segmentation (which expects to know classes of interest), nor retrain a model from scratch. Instead, we propose to guide CLIP prediction with an unsupervised foreground/background segmentation method, which to the best of our knowledge has not yet been explored. 

\section{\NAME}
We tackle the problem of open-vocabulary semantic segmentation with no supervision. Let us consider a set of queries $t_j \in T$ formulated in natural language. 
Our goal is to localize each query if present in the image, yielding one mask per query, i.e. a segmentation map. Our approach, summarized in Fig.~\ref{fig:overview}, consists of two stages. In Sec.~\ref{sec:clippool}, we describe our first step, where we run our proposed multi-scale dense inference to obtain coarse semantic maps by running CLIP on image patches at different scales. Our second step, which we cover in Sec.~\ref{sec:obj_fusion}, consists of refining the initial segmentation using an off-the-shelf foreground-background extractor. 

\subsection{Dense inference with CLIP}
\label{sec:clippool}

\begin{figure}[tp]
    \centering
    \begin{tabular}{ccc}
\includegraphics[width=0.285\linewidth]{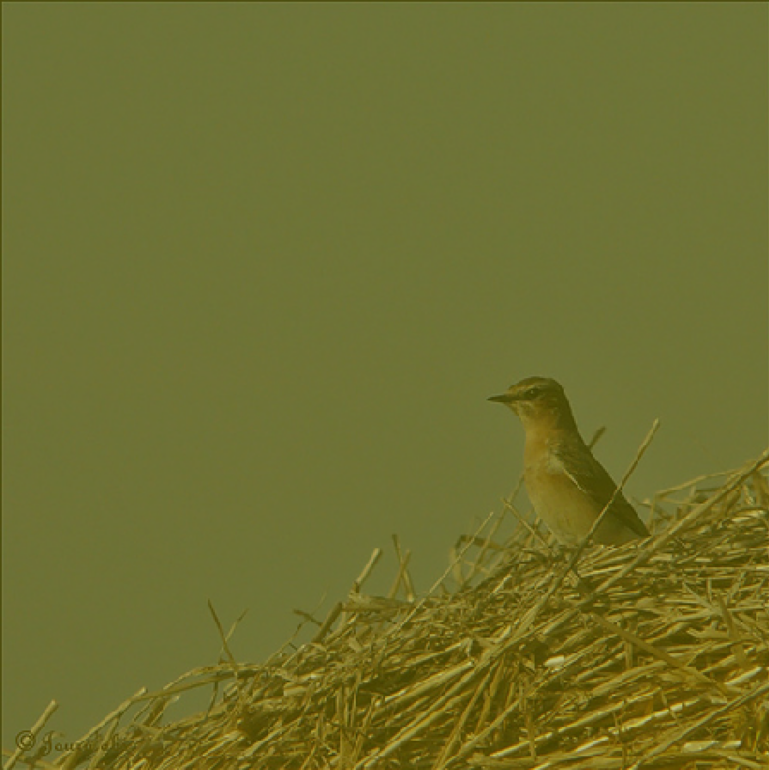}  & 
\includegraphics[width=0.285\linewidth]{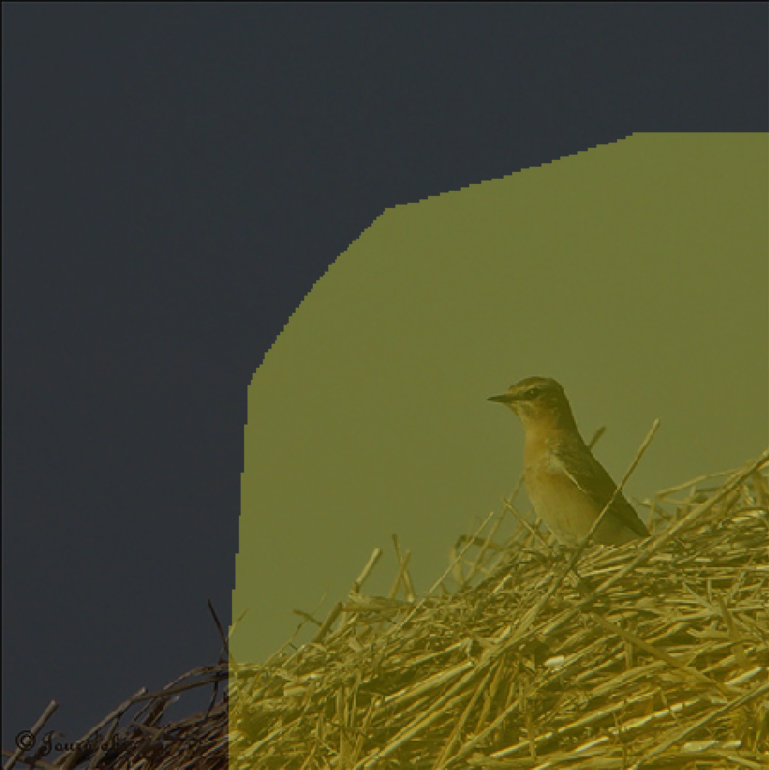} 
& 
\includegraphics[width=0.285\linewidth]{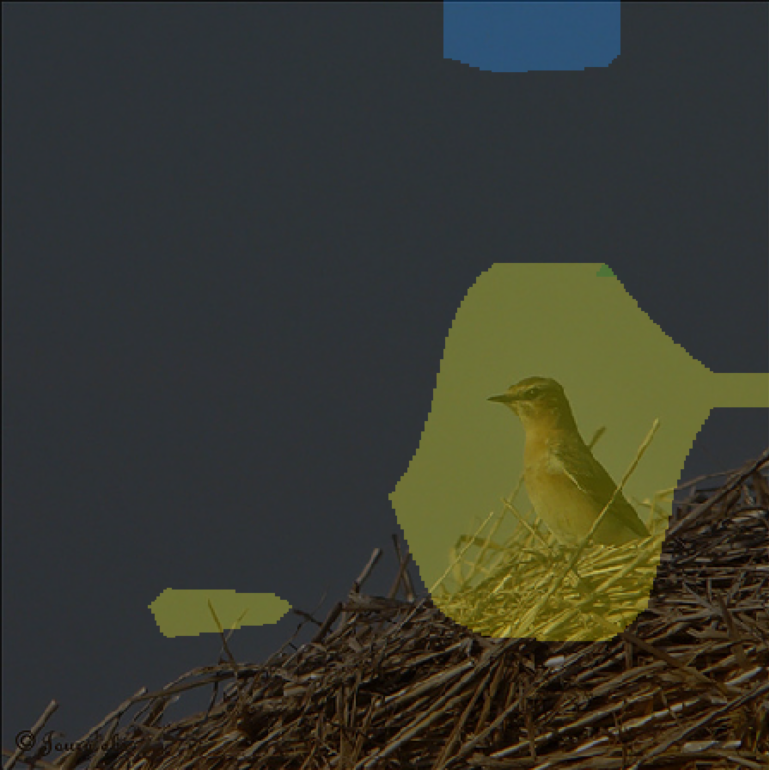}
\\
scale $0$ & scale $1$ & scale $2$ \\
\includegraphics[width=0.285\linewidth]{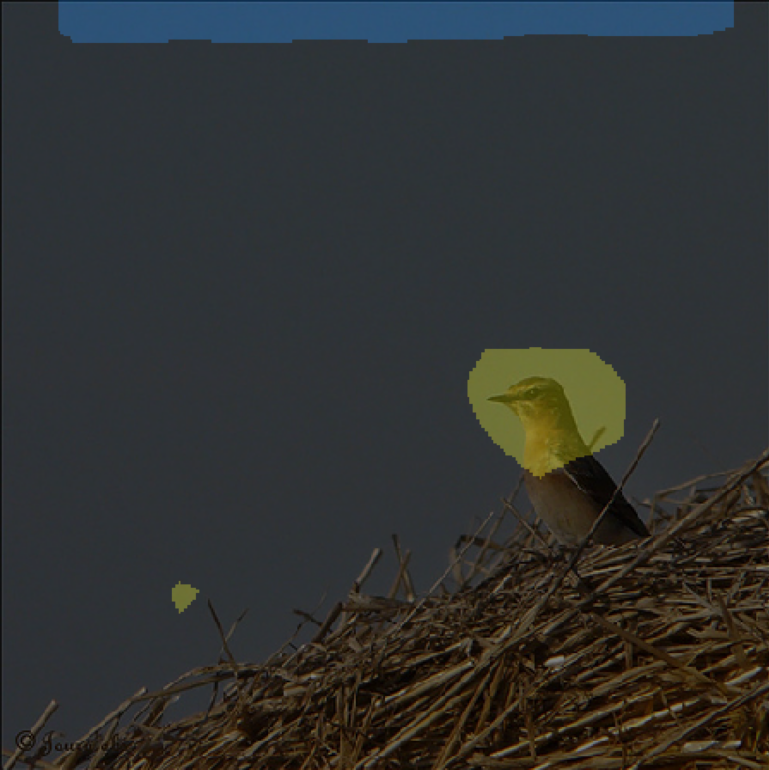}  & 
\includegraphics[width=0.285\linewidth]{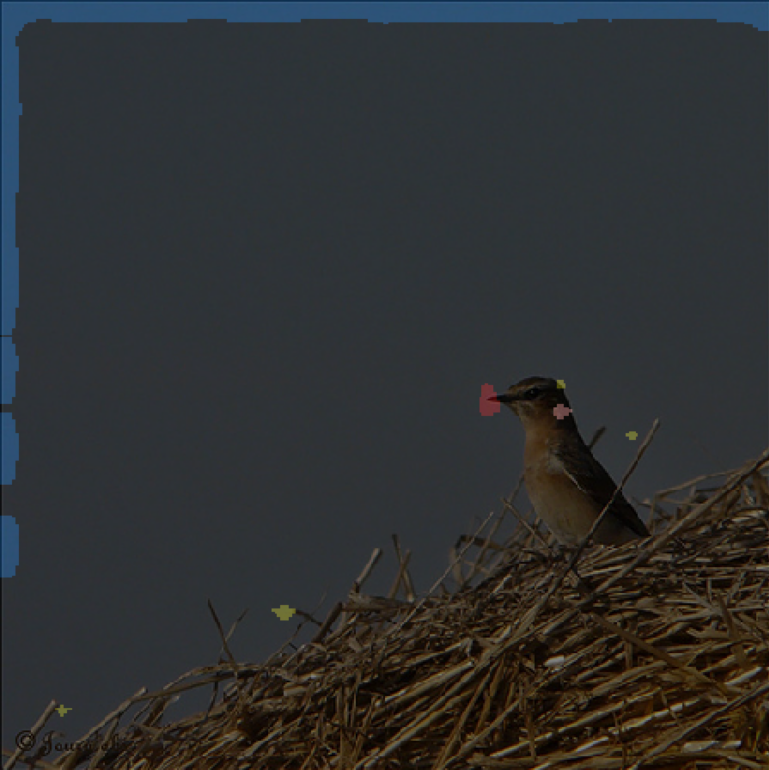} 

& 
\includegraphics[width=0.285\linewidth]{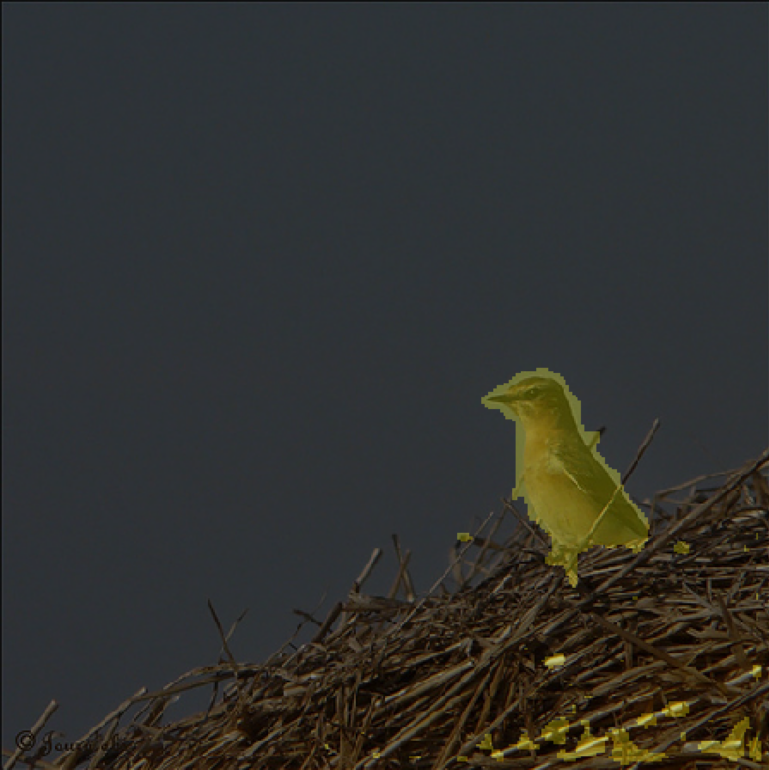}\\
scale $3$ & scale $4$ & \NAME \\

    \end{tabular}
    \caption{\textbf{Multi-scale maps $\clipmap_s$ of \NAME.} Our method produces multi-scale predictions of dense logits that capture information at different levels of granularity. While the coarsest $(s=0)$ scale 
    pools global information 
    from the image, finer scales induce more localized and potentially discover different object classes. While our method can incorporate as many scales as needed, very fine resolutions such as scale $4$ and above do not generate meaningful information.}
    \vspace{-0.3cm}

    \label{fig:intermediate_products}
\end{figure}

Our method leverages CLIP~\cite{radford2021learning} as a backbone. Contrary to most CLIP-based approaches for zero-shot semantic segmentation (as discussed in Sec.~\ref{ssec:related_zeroshot}), we do not rely on patch tokens within the image encoder. Instead, we leverage CLIP's zero-shot capabilities by running the model densely on image partitions. Thus, we calculate the alignment between each of the multi-scale patches and the considered textual queries.

\paragraph{Encoding prompts}
Given a set of textual queries 
$\queries$, we encode a prompt $t_j \in \queries$ with $j \in [0, |\queries|]$  using CLIP text encoder $\cliptxt(t_j) \in \mathbb{R}^d$, where $d=512$ in CLIP. Following previous works~\cite{cha2022tcl}, we formulate a prompt as: \texttt{a photo of a \{\textbf{$t_j$}\}}. In what follows we use the notation $\cliptxt(\queries) \in \mathbb{R}^{d \times |\queries|}$ as the set of encodings of all text queries $\queries$. Note that we always consider a \texttt{background} class, thus we always assume $t_0 = $\text{ \texttt{background}}.

\paragraph{Multi-scale image partitioning} Given an input image $x \in \mathbb{R}^{H \times W \times 3}$, we first reshape it into a sequence of 2D patches $\partition=\{x_i \in \mathbb{R}^{P \times P \times 3}\}_{i=1..N}$, where $N = \lceil \frac{H}{P}\rceil \cdot \lceil\frac{W}{P}\rceil$ and $P\times P$ is the patch size. We then extract visual embeddings $\clipimg(x_i) \in \mathbb{R}^d$ for each patch. 
In practice, since we use a ViT~\cite{dosovitskiy2020image} as our visual encoder $\clipimg$, and take the [CLS] output as our visual embedding $\clipimg(x_i)$\footnote{For convolutional network backbones such as Resnet~\cite{he2016deep},  we could use the output of AvgPool.}. We can then run the same partitioning for different scales $\scales$ using different patch sizes, yielding a set of partitions $\{\partition_s\}_{s\in\scales}$.

\paragraph{Dense similarities extraction}

Given our multi-scale partitions, and a text query $t\in \queries$, we build a dense similarity map $\clipmap_s^t$ for each scale $s\in \scales$, such that:

\begin{equation}
    \clipmap_s^t = \upsample\limits_{H,W}\left(\bigcup\limits_{x \in \partition_s} \left[\clipimg(x) \otimes \cliptxt(t)\right]\right), 
\end{equation}

where $\bigcup$ is a merging operator that puts patches back onto a 2D grid, $\otimes$ denotes the inner product computed between the visual and text embeddings and $\upsample\limits_{H, W}$ is a bilinear up-sampling operator which upsamples its input to the resolution $H\times W$.

The resulting map $\clipmap_s^t \in \mathbb{R}^{H\times W}$ yields an estimate of the similarity between each pixel in the input image and a text query $t \in \queries$. In Fig.~\ref{fig:intermediate_products} we show aggregated similarity maps $\clipmap_s = \{\clipmap_s^t\}_{t=1..|\queries|} \in \mathbb{R}^{H\times W\times |\queries|}$ obtained at different scales. We can see that while the coarsest scale $s=0$ is responsible for pooling global information on the objects to segment, finer scales $s>1$ result in more localized maps. 

Having obtained similarity maps $\clipmap_s^t$ for each text query $t\in \queries$ and each scale $s\in\scales$
we aggregate the multi-scale predictions into a map $\clipmask^t$ such that:

\begin{equation}
\clipmask^t = \dfrac{1}{|S|}\sum\limits_{s\in \scales}\clipmap_s^t.
\label{eq:multiscale_map}
\end{equation}

\subsection{Guided segmentation}
\label{sec:obj_fusion}
Finally, we propose to refine the multi-scale segmentation maps $\clipmask^t$ of Eq.~\ref{eq:multiscale_map} using an objectness map produced by an
off-the-shelf \emph{unsupervised} foreground-background segmentation method~\cite{shin2022selfmask,simeoni2023found}, noted $\Theta$. As discussed in related work, such methods exploit self-supervised features, e.g.~\cite{caron2021dino}, to discover the pixels likely depicting objects.

When fed with an image $x \in \mathbb{R}^{H\times W}$, the foreground-background segmentation method produces an output $\Theta(x)\in [0,1]^{H\times W}$ with a per-pixel confidence score close to $1$ 
for a pixel in a foreground object. We use this proxy for objectness estimation to refine our segmentation masks $\clipmask^t$.
In particular, we refine $\clipmask^t$ using the output of $\Theta$ for each text query $t\in \queries$ except for $t_0=\text{\texttt{background}}$, 
where we take the complement of $\Theta(x)$ following:
\begin{equation}
    \saliency^t =\begin{cases}
    1 - \Theta(x) & \text{if $t=\text{\texttt{background}}$} , \\
    \Theta(x) & \text{otherwise},
    \end{cases}
\end{equation}

such that similarities with the background class are down-weighted for pixels deemed as \emph{salient} by $\Theta$. 

Finally, we compute the output of \NAME as:

\begin{equation}
    M = \softmax\limits_{t\in \queries} \left(\clipmask^t \odot \saliency^t\right) \in \mathbb{R}^{H\times W \times |\queries|},
\end{equation}

where $\odot$ denotes the Hadamard product and $\softmax$ is the softmax operator computed over text queries. 
In Fig.~\ref{fig:intermediate_products} we show how the aggregation of all scales paired with the guidance of $\Theta$ results in accurate object segmentation.

\section{Experiments}
In this section, we present the experiments conducted to evaluate our method and justify particular design choices. First, in Sec~\ref{ssec:exp_setup} we give details about our experimental setup.
In Sec.~\ref{ssec:exp_comparison}, we discuss how our method compares against other open-vocabulary semantic segmentation approaches, both quantitatively and qualitatively. We then give more insight into our method with a series of ablations (Sec.~\ref{ssec:ablation}), failure mode analysis (Sec.~\ref{ssec:failure_cases}) and finally real open-world evaluation (Sec.~\ref{ssec:in_the_wild}).

\newlength{\pascalwidth}
\setlength{\pascalwidth}{0.16\textwidth}
\newlength{\cocowidth}
\setlength{\cocowidth}{0.15\textwidth}
\newlength{\resultsheight}
\setlength{\resultsheight}{1.93cm}

\newcommand{\pascalresult}[2]{%
\begin{minipage}[b]{\pascalwidth}
     \centering
     \includegraphics[width=\textwidth, height=\resultsheight]{figures/images/result_samples/pascal_voc/#1_#2.png}
\end{minipage}}
\newcommand{\cocoresult}[2]{%
\begin{minipage}[b]{\cocowidth}
     \centering
     \includegraphics[width=\textwidth, height=\resultsheight]{figures/images/result_samples/coco_#1_#2.png}
\end{minipage}}
\newcommand{\pascalresultjpeg}[2]{%
\begin{minipage}[b]{\pascalwidth}
     \centering
     \includegraphics[width=\textwidth, height=\resultsheight]{figures/images/result_samples/pascal_voc/#1_#2.jpg}
\end{minipage}}
\newcommand{\cocoresultjpeg}[2]{%
\begin{minipage}[b]{\cocowidth}
     \centering
     \includegraphics[width=\textwidth, height=\resultsheight]{figures/images/result_samples/coco_#1_#2.jpg}
\end{minipage}}

\begin{figure*}[t]
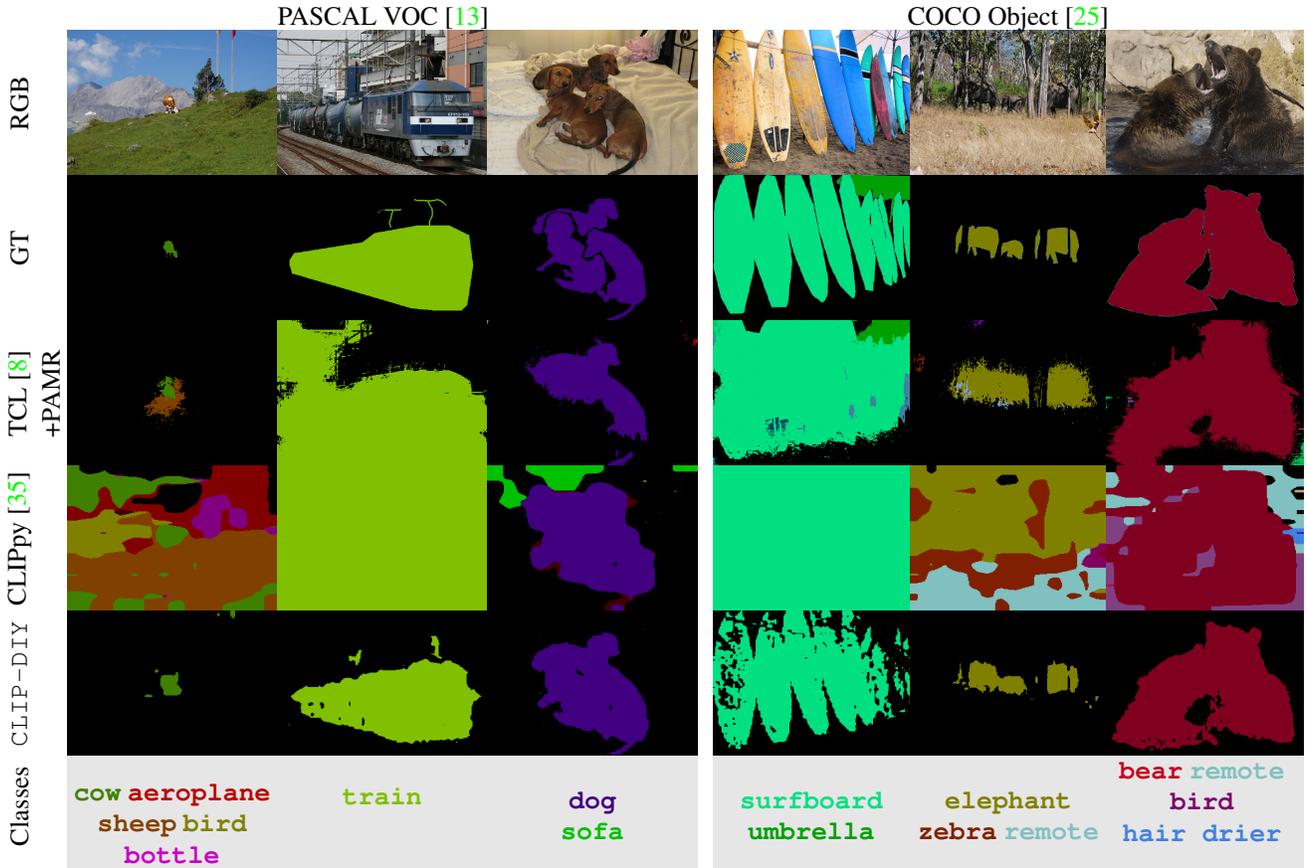

\centering
\renewcommand{\arraystretch}{0}
\setlength\tabcolsep{0pt} 

\begin{center}
\begin{tabular}{
>{\centering\arraybackslash}m{1.2em} 
>{\centering\arraybackslash}m{1.2em} 
>{\centering\arraybackslash}m{\pascalwidth}
>{\centering\arraybackslash}m{\pascalwidth}
>{\centering\arraybackslash}m{\pascalwidth}@{\hskip 0.08in}
>{\centering\arraybackslash}m{\cocowidth}
>{\centering\arraybackslash}m{\cocowidth}
>{\centering\arraybackslash}m{\cocowidth}
}
~ & ~ & \multicolumn{3}{c}{PASCAL VOC~\cite{pascal-voc-2012}}~ & \multicolumn{3}{c}{COCO Object~\cite{lin2014microsoft}}\\
\rotatebox{90}{RGB} & & 
\pascalresult{pascal_2010_002701}{img} &
\pascalresult{pascal_2009_001333}{img} &
\pascalresultjpeg{pascal_2007_001239/pascal_2007_001239}{img} & 
\cocoresult{127517}{img} &
\cocoresult{126216}{img} & 
\cocoresult{127955}{img} \\

\rotatebox{90}{GT} & &
\pascalresult{pascal_2010_002701}{gt} &
\pascalresult{pascal_2009_001333}{gt} &
\pascalresult{pascal_2007_001239/pascal_2007_001239}{gt} & 
\cocoresult{127517}{gt} &
\cocoresult{126216}{gt} &
\cocoresult{127955}{gt} \\


\rotatebox{90}{TCL~\cite{cha2022tcl}}  & \rotatebox{90}{+PAMR} & 
\pascalresult{pascal_2010_002701}{tcl} &
\pascalresult{pascal_2009_001333}{tcl} &
\pascalresult{pascal_2007_001239}{tcl} & 
\cocoresult{127517}{tcl} &

\cocoresult{126216}{tcl} &

\cocoresult{127955}{tcl} \\

\rotatebox{90}{CLIPpy~\cite{clippy2022}} & &
\pascalresult{pascal_2010_002701}{clippy} &
\pascalresult{pascal_2009_001333}{clippy} &
\pascalresult{pascal_2007_001239}{clippy} & 
\cocoresult{127517}{clippy} &
\cocoresult{126216}{clippy} & 
\cocoresult{127955}{clippy} \\

\rotatebox{90}{\NAME} & &
\pascalresult{pascal_2010_002701}{pred_final} &
\pascalresult{pascal_2009_001333}{pred_final} &
\pascalresult{pascal_2007_001239/pascal_2007_001239}{pred_final} & 
\cocoresult{127517}{pred} &
\cocoresult{126216}{pred_final} &
\cocoresult{127955}{pred_final} \\

\vspace{1pt}
 \rotatebox{90}{Classes} & ~ & 
 {\cellcolor[gray]{.9} \makecell{\promptstyle{cow}{cow} \promptstyle{aeroplane}{aeroplane} \\ \promptstyle{sheep}{sheep} \promptstyle{bird_pascal}{bird} \\
 \promptstyle{bottle}{bottle}
}}
&
{\cellcolor[gray]{.9}\promptstyle{train}{train}}
&
 {\cellcolor[gray]{.9}\makecell
{\promptstyle{dog}{dog} \\ \promptstyle{sofa}{sofa} }}
&
{\cellcolor[gray]{.9}\makecell 
{\promptstyle{surf}{surfboard} \\ \promptstyle{umbrella}{umbrella}}}
& 
{\cellcolor[gray]{.9}\makecell{\promptstyle{elephant}{elephant}  \\ \promptstyle{zebra}{zebra} \promptstyle{remote}{remote} }}
& 
{\cellcolor[gray]{.9} \makecell{\promptstyle{bear}{bear} \promptstyle{remote}{remote} \\ \promptstyle{bird_coco}{bird} \\
\promptstyle{hair}{hair drier}} }

\end{tabular}
\end{center}
    \vspace{-1.5em}
    \caption{   
    \textbf{Qualitative open-vocabulary segmentation results}. We compare our method against CLIPpy~\cite{clippy2022} and TCL (with PAMR post-processing)~\cite{cha2022tcl}. Our method consistently outperforms two other methods by producing accurate segmentation masks. TCL and CLIPpy also both suffer from hallucinating 
    classes based on context, such as the \promptstyle{aeroplane}{aeroplane} and \promptstyle{sheep}{sheep} in 1st column, or \promptstyle{zebra}{zebra} in 5th column. All pixels annotated in black are from the \promptstyle{black}{background} class.
    }
    \vspace{-0.1cm}
    \label{fig:results_main}
\end{figure*} 

\subsection{Experimental setup}
\label{ssec:exp_setup}

\paragraph{Datasets \& metric}
We evaluate our method on two common semantic segmentation benchmarks: PASCAL VOC 2012~\cite{pascal-voc-2012}
and COCO~\cite{lin2014microsoft}, comprising of 20 
and 80 foreground classes respectively. PASCAL VOC has an additional \texttt{background} class, and we adopt a unified protocol~\cite{xu2022groupvit, cha2022tcl} considering a \texttt{background} class in all datasets.
We evaluate results with the mean Intersection-over-Union (mIoU) metric. 
For evaluation, we resize input images to have the shorter side of length 448 following~\cite{cha2022tcl}.

\paragraph{Implementation details}
If not otherwise specified, we use the CLIP ViT-B/32 model OpenCLIP version~\cite{openclip} trained with LAION~\cite{schuhmann2022laionb}. The input images are resized to $224 \times 224$ and the patch size is $32 \times 32$. We empirically find that running our model on 3 different scales i.e. $|\scales|=3$ with patch sizes of $P_0=256, P_1=128, P_2=64$ gives the best results for both evaluated datasets. We discuss this later in Sec. \ref{ssec:ablation}.

\newlength{\backbonelength}
\setlength{\backbonelength}{2.4cm}
\begin{table*}[htbp]
\centering 
\begin{tabular}{l@{\hskip 0.2cm}c
>{\centering\arraybackslash}m{\backbonelength}
>{\centering\arraybackslash}m{\backbonelength}
@{\hskip 0.5cm}cc@{}}
\toprule
\textbf{Method} & extra training ? & \multicolumn{2}{c}{Backbones} & \textbf{PASCAL} &  \textbf{COCO} \\
~ & ~ & Visual & Text & \textbf{VOC} & \textbf{Object} \\
\midrule

ReCo$^{\dagger}$~\cite{shin2022reco} & \checkmark & ViT-L/14* & CLIP-ViT-L/14* & 25.1  &  15.7 \\

ViL-Seg~\cite{liu2022open} & \checkmark & ViT-B/16  &  & 37.3 & -  \\
MaskCLIP+$^{\dagger}$~\cite{zhou2022maskclip} & \checkmark & ResNet101~\cite{he2016deep} &  & 38.8 &  20.6 \\

CLIPpy~\cite{clippy2022} & \checkmark & ViT-B/16 &  T-5 \cite{raffel2020exploring} & 52.2 &  \textbf{32.0} \\
GroupViT~\cite{xu2022groupvit} & \checkmark & ViT-S/16 & 12T & 52.3 &  - \\
ViewCo~\cite{ren2023viewco}  & \checkmark & ViT-S/16 & 12T & 52.4 & 23.5 \\
SegCLIP~\cite{lou2022segclip} & \checkmark &  ViT-B/16  & CLIP-ViT-B/16 & 52.6 &  26.5   \\
OVSegmentor~\cite{xu2023learning} & \checkmark  & ViT-B/16 &  BERT-ViT-B/16  & 53.8 &  25.1 \\
TCL~\cite{cha2022tcl} {\footnotesize+ PAMR~\cite{pamr}} & \checkmark & ViT-B/16  & CLIP-ViT-B/16 & \underline{55.0} & \underline{31.6} \\
\NAME (ours) &  & ViT-B/16 &  CLIP-ViT-B/16 & 
59.0  &  30.4 \\
\NAME (ours) &  &  ViT-B/32 &  CLIP-ViT-B/32 & \textbf{59.9} & 31.0\\

\bottomrule
\end{tabular}
\caption{\textbf{Zero-shot open-vocabulary segmentation}. Comparison of our approach to the state of the art (under the mIoU metric). While our method does not need any additional training it performs significantly better than the current SOTA on PASCAL VOC \textbf{(+4.9)} and performs on par with its competitors on COCO object, ranking 3rd on COCO. We mark with  $^{\dagger}$ results from~\cite{cha2022tcl}. All methods are evaluated considering that \texttt{background} is a class of the dataset. We note with $*$ when more than one backbone was used, we refer here to CLIP-like backbones. GroupViT and ViewCo use a 12 Transformer layers backbone following \cite{radford2021learning}, noted 12T.}
\label{tab:main_results}
\end{table*}

\paragraph{Baselines} We compare our method with existing 
state-of-the-art zero-shot open-vocabulary methods.
In particular, we evaluate against methods including 
self-trained MaskCLIP+~\cite{zhou2022maskclip}, learning grouping strategies: GroupViT~\cite{xu2022groupvit}, SegCLIP~\cite{lou2022segclip}, ViL-Seg~\cite{liu2022open}, OVSegmentor~\cite{xu2023learning}, ViewCo~\cite{ren2023viewco} , using class prototypes: ReCo$^{\dagger}$~\cite{shin2022reco}, text-grounding strategy: TCL~\cite{cha2022tcl} and with CLIPpy \cite{clippy2022} which uses a T-5 backbone and improves dense abilities of CLIP.
\footnote{We do not compare against~\cite{mukhoti2023pacl} due to lack of comparable results and open source code.}
We detail in Tab.~\ref{tab:main_results} 
the VLM backbones used per method and the
if additional training data was. 
Every method (except for MaskCLIP) requires training a specific module/model used to get denser predictions; instead, we use the vanila CLIP model. 

\paragraph{A note on fair comparison} Following TCL~\cite{cha2022tcl}, we use a unified evaluation protocol corresponding to an open-world scenario
where prior access to the target data before evaluation is not
allowed. In particular, we do not consider query expansion, e.g. class name expansion or rephrasing. As discussed in~\cite{cha2022tcl} exploring language biases can greatly improve the overall segmentation performance. However, we only use the original class names from the compared datasets. We also report best-reported scores for all methods. It is to be noted that TCL uses a post-processing technique, namely PAMR~\cite{pamr} while other methods do not.

\subsection{Results}
\label{ssec:exp_comparison}
In this section, we compare our method to previous work both quantitatively and qualitatively.

\paragraph{Quantitative results} We summarize in Tab.~\ref{tab:main_results}, the comparison of our \NAME to baselines. We report results of \NAME with both CLIP ViT-B/32 and CLIP ViT-B/16, given that most methods use the latter.

First, comparing our results with the two different backbones, we remark that better scores are obtained with CLIP ViT-B/32 which patch inputs are larger, and are therefore less expensive at inference time.
We believe this could be explained by an existing upper bound on CLIP accuracy w.r.t the level of granularity; patches too small might induce noisy classification---as we observed in Fig.~\ref{fig:intermediate_products}. 

Compared to baselines, we observe that our method achieves the best mIoU result on PASCAL VOC dataset and outperforms all previous works by more than 4 mIoU pts, and such without post-processing. This result is particularly interesting given that our method does not require dedicated training to improve CLIP segmentation abilities but instead leverages the unsupervised object localization method.

Moreover, we obtain $31.0$ mIoU on COCO when the best-performing method CLIPpy achieves $32.0$, so just 1 mIoU pt better than our approach. We can observe in Fig.\ref{fig:results_main} that CLIPpy discovers more queries per image than us, even though its segmentation outputs appear to be noisier. Such results might suit a better COCO benchmark and less PASCAL VOC.

\paragraph{Qualitative results} We qualitatively compare here our method with best performing TCL~\cite{cha2022tcl} and CLIPpy~\cite{clippy2022} in Fig.~\ref{fig:results_main}.
Our method consistently produces better masks with better object boundaries, which we attribute to the high-quality saliency maps. Moreover, we observe that \NAME produces correct semantic results on the foreground objects, with fewer artefacts than CLIPpy. Our method seems also less sensitive to biases, for instance, both TCL and CLIPpy hallucinate the \texttt{sheep} class on the grass (on the very left image) and CLIPpy also predicts \texttt{aeroplane} in the sky (in most left image) and \texttt{zebra} next to the elephants (middle image in COCO dataset). 

\subsection{Ablations}
\label{ssec:ablation}

\begin{table}[t]
\centering 
  \begin{tabular}{lccc}
\toprule
    Method&\ PASCAL VOC & COCO \\
    \midrule
    \NAME & 59.9 & 31.0 \\
    w/o multi-scale & 56.0 & 25.9 \\
    w/o objectness & 24.1 & 15.5 \\

    \bottomrule
\end{tabular}
\caption{\textbf{Ablation studies.} We find that while all components are improving the performance of our method, objectness is particularly critical. Numbers are reported using a ViT-B/32 backbone. }
\vspace{-0.4cm}
\label{tab:ablations}
\end{table}

In this section, we perform ablation studies to validate the individual choices in the design of \NAME. 

First, we study in Tab.~\ref{tab:ablations} the impact of the different elements of our method. In particular, we investigate the impact of using a multi-scale mechanism and leveraging the objectness produced by the foreground/background segmenter. We notice that by removing our multi-scale scheme we drop results by 3.9 and 5.1 mIou pts on Pascal and COCO respectively, showing the benefit of considering patches of different sizes. Additionally, the largest drop is observed when removing the foreground/background saliency guidance showing the effectiveness of combining CLIP with the current lightest unsupervised object localization model, FOUND.

\paragraph{Multi-scale approach} 

\begin{table}[t]
    \centering
    \begin{tabular}{ccccc|cc}
    \toprule
       
       \multicolumn{5}{c|}{Scales used} & PASCAL & COCO \\
       0  & 1 & 2  & 3 & 4  & VOC & Object \\
       \cmidrule(l{2pt}r{2pt}){1-5} \cmidrule(l{2pt}r{2pt}){6-7} 
       \checkmark &  &  &  &  & 56.0 & 25.9 \\
       \checkmark & \checkmark &  &  &  & 58.8 & 28.7 \\
       \checkmark & \checkmark & \checkmark &  &  & \textbf{59.9} & \textbf{31.0} \\
       \checkmark & \checkmark & \checkmark & \checkmark &  & 59.5 & 29.9 \\
       \checkmark & \checkmark & \checkmark & \checkmark & \checkmark & 59.3 & 29.4 \\
        & \checkmark & \checkmark & \checkmark &  & 53.8 & 26.4 \\
       \bottomrule
    \end{tabular}
    \caption{\textbf{Ablation of our multi-scale approach.} To validate our multi-scale design, we report the results of our method with different sets of scales, by progressively adding more fine-grained scales. We find empirically that after scale 3, adding more scales gives similar or worse results. We also report results without the first scale $s=0$ (global scale), which leads to worse results. Numbers are reported using a ViT-B/32 backbone.
    }
    \label{tab:ablation_scales}
\end{table}

We conduct an ablation study on the scales used in our multi-scale scheme to generate the predictions.
We progressively add inner scales in Eq.~\ref{eq:multiscale_map} and report the resulting accuracy on all datasets. The results, detailed in Tab.~\ref{tab:ablation_scales} show an optimum when using three fine scales. Adding more does not appear to improve results nor downgrade them, showing the stability of our method.
Interestingly, we also see that removing information from the global scale greatly reduces performance on all datasets (-5.7/-3.5 mIoU pts on PASCAL VOC/COCO respectively). In Fig.~\ref{fig:qualitative_ablation} we visualize the individual contribution of each scale to our final prediction. As in Fig.~\ref{fig:intermediate_products}, we observe that coarse scales capture the global context, while finer scales capture more local one, such that objects can be separated.

\newlength{\quallength}
\setlength{\quallength}{0.4\linewidth}
\begin{figure}[t]
\centering
\begin{tabular}{cc}
 \includegraphics[width=\quallength]{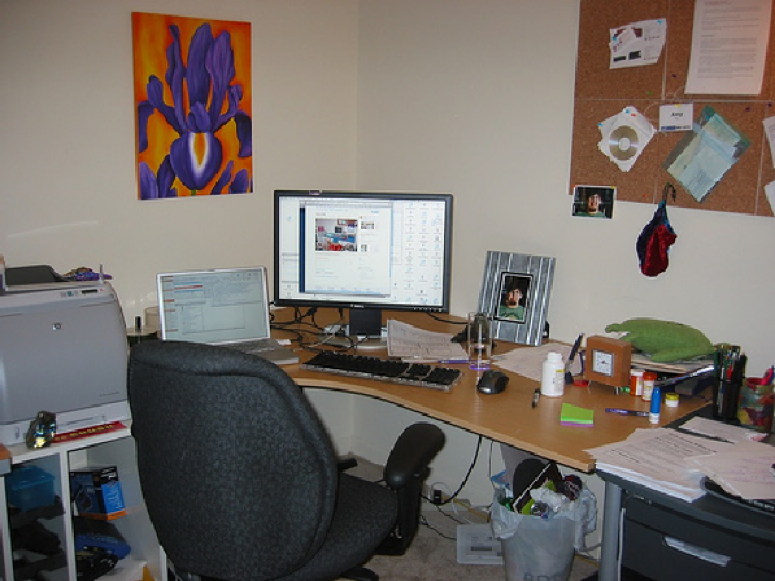}  & 
\includegraphics[width=\quallength]{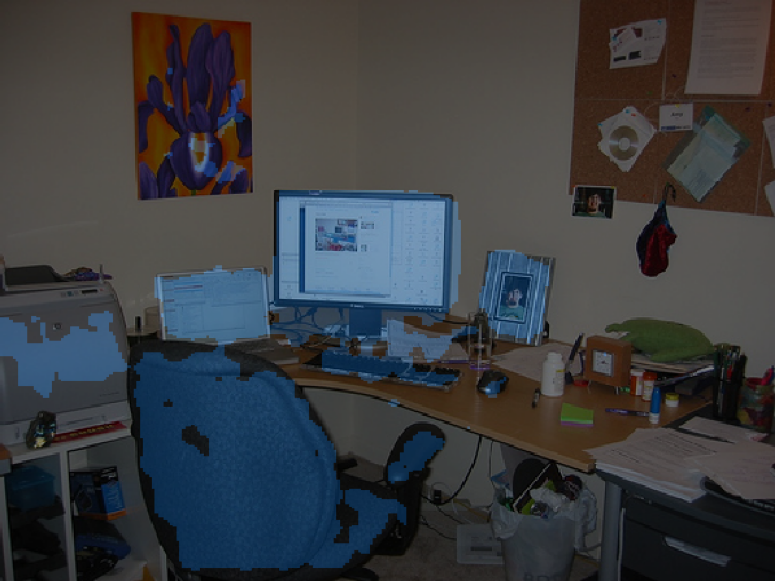} 
\\
RGB & scale 0 \\
\includegraphics[width=\quallength]{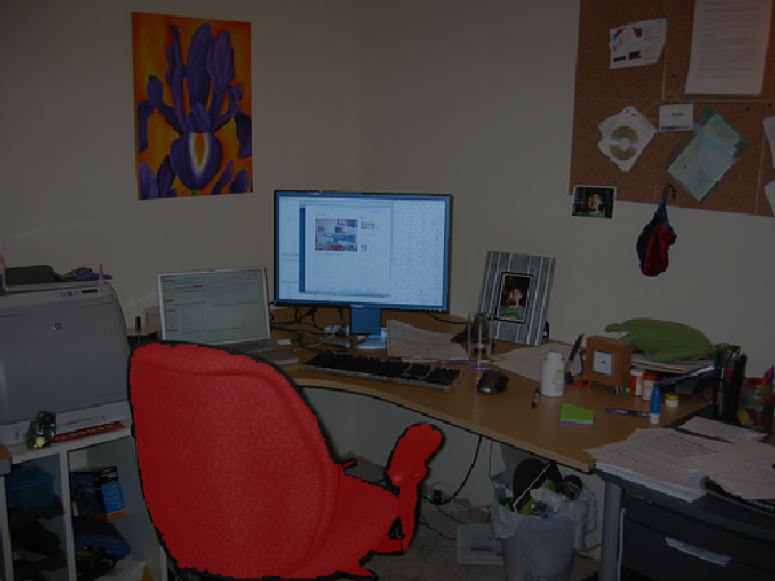}
&
\includegraphics[width=\quallength]{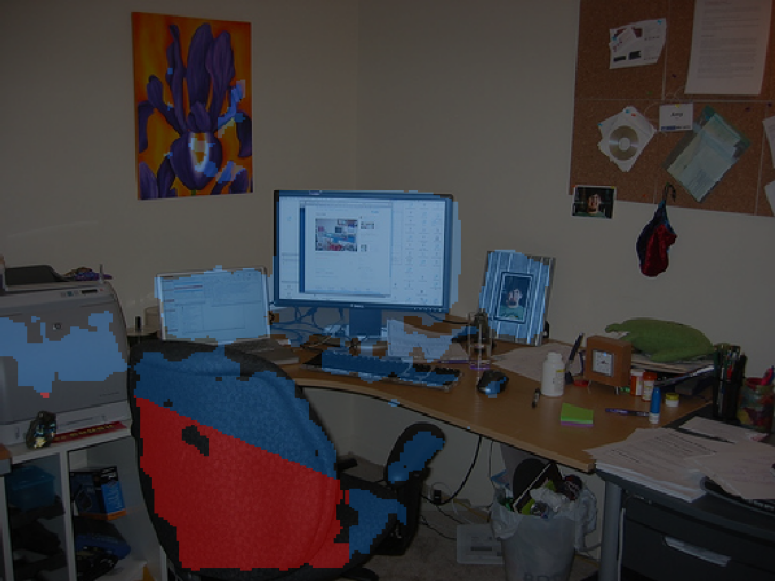}
\\
GT & scale 1 \\
\includegraphics[width=\quallength]{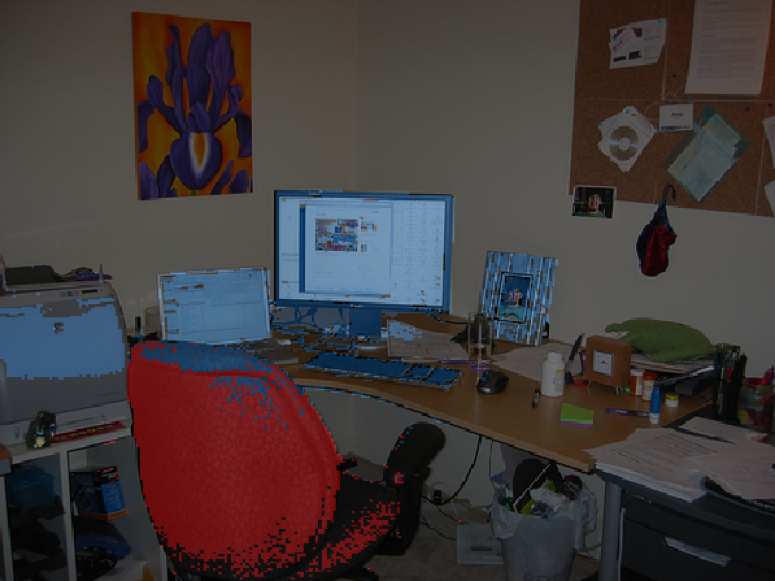} 
&
\includegraphics[width=\quallength]{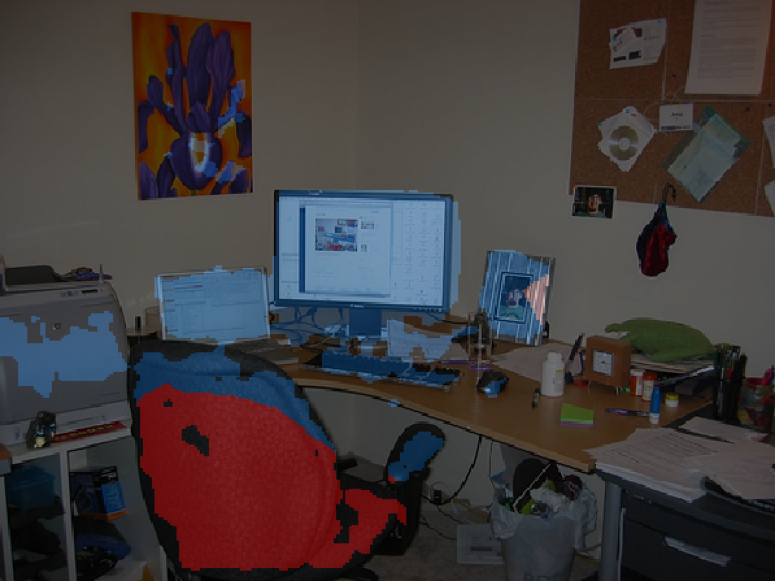} \\
\NAME 
& scale 2 \\
    {\cellcolor[gray]{1}
        \promptstyle{chair}{chair}
    }
    &
        {\cellcolor[gray]{1}
        \promptstyle{tv/monitor}{tv/monitor}
    }
    \end{tabular}
    \vspace{-0.3cm}
    \caption{\textbf{Qualitative ablation study.} In the right column we show the results when progressively adding more scales in the multi-scale approach. Running with 3 scales enables segmenting most of the chair.}
    \vspace{-0.4cm}
    \label{fig:qualitative_ablation}
\end{figure}

\begin{table}[t]
\centering 
  \begin{tabular}{lcc}
\toprule
Saliency Method & w. training & PASCAL VOC \\
\midrule
     FOUND-bkg~\cite{simeoni2023found} & \xmark & 48.4 \\
    FOUND~\cite{simeoni2023found} & \checkmark & 59.9 \\ 
    CutLER saliency~\cite{wang2023cut} & \checkmark & 55.4 \\
    CutLER mask~\cite{wang2023cut} & \checkmark & 50.8 \\
    \bottomrule
\end{tabular}
\caption{\textbf{Comparison of different objectness methods} used for objectness-guided fusion. We find that the version of FOUND that was trained \emph{in the original paper} performs the best, and therefore keep this method as our objectness guide. }

\vspace{-0.3cm}
\label{tab:saliency}
\end{table}

\paragraph{Foreground segmenters}

We compare different foreground-background segmenters and the overall performance of our method using each one of them. The results, summarized in Tab.~\ref{tab:saliency}, show that our method performs the best when using FOUND, more specifically the FOUND model that has been re-trained with self-training in the original work. We also experiment with CutLER~\cite{wang2023cut}, which performs unsupervised instance segmentation. We use the predicted instance masks or compute a saliency. In both cases, we obtain slightly worse results. We give more details in Sec.~\ref{ssec:unsupervised_obj} of Supplementary material.

\subsection{Failure cases}
\label{ssec:failure_cases}
\paragraph{Failure cases} We qualitatively analyse failure cases of our method by showing a couple of examples in Fig.~\ref{fig:failure_cases}.
We observe that some of the failures are due to inaccurate annotations: in (a) \texttt{bear} is only partially annotated and in (b) a mask for an \texttt{elephant} is annotated too coarsely. Our method, benefiting from FOUND's accurate saliency predictions is able to produce better segmentation masks. We also observe that our method is limited by the quality of the saliency (c, d), which we comment more on below.

\paragraph{Ambiguities} 
In the examples (c) and (d) of Fig.~\ref{fig:failure_cases}, we observe that our method can fail to segment objects with significant overlap, resulting in ambiguity regarding the foreground class, which is especially harmful when annotations are too coarse or incomplete. In (c), only the \texttt{bench} is annotated, while our method segments only the orange, which would result in a low IoU score. In (d), the opposite behaviour occurs, where \NAME discovers more classes than the annotated ones. Finally, similarly to most of the open-vocabulary methods based on CLIP, our method suffers from sensitivity to text ambiguities. We show more failure cases in Sec. ~\ref{sec:failures_sup} of the Supplementary material.

\newcommand{\cocoresulttwo}[2]{%
\begin{minipage}[b]{0.99\linewidth}
     \centering
     \includegraphics[width=\textwidth, height=2.6cm]{figures/images/result_samples/#1_#2.png}
\end{minipage}}
\newcommand{\cocoresulttwojpeg}[2]{%
\begin{minipage}[b]{0.99\linewidth}
     \centering
     \includegraphics[width=\textwidth, height=2.6cm]{figures/images/result_samples/#1_#2.jpg}
\end{minipage}}

\begin{figure}[htbp]
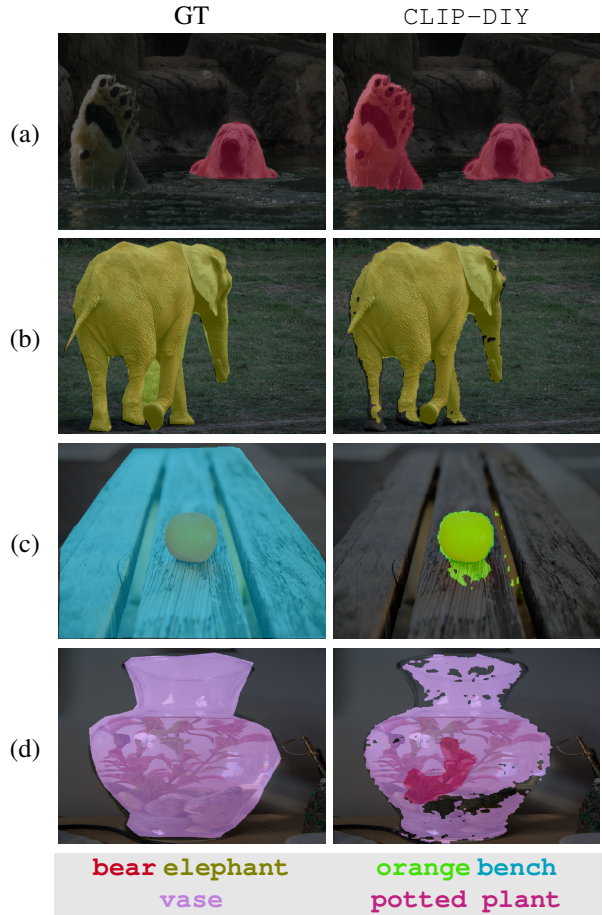

\centering
\renewcommand{\arraystretch}{1} 
\setlength\tabcolsep{1pt}  
\begin{center}
\begin{tabular}{
>{\centering\arraybackslash}m{2em}
>{\centering\arraybackslash}m{0.43\linewidth}
>{\centering\arraybackslash}m{0.43\linewidth}}
~ & GT & \NAME \\
(a) &
\cocoresulttwo{coco_85478}{gt_overlay} &
\cocoresulttwo{coco_85478}{final_pred_overlay} \\

(b) &
\cocoresulttwo{coco_155571}{gt_overlay} &
\cocoresulttwo{coco_155571}{final_pred_overlay} \\

(c) &

\cocoresulttwo{coco_446117}{gt_overlay}  & 
\cocoresulttwo{coco_446117}{final_pred_overlay} \\

(d) &
\cocoresulttwo{coco_298904}{gt_overlay} &
\cocoresulttwo{coco_298904}{final_pred_overlay} \\
&  {\cellcolor[gray]{.9}\makecell
{\promptstyle{bear}{bear} \promptstyle{elephant}{elephant} \\ \promptstyle{vase}{vase}}}
&  {\cellcolor[gray]{.9}\makecell
{\promptstyle{orange}{orange} \promptstyle{bench}{bench} \\ \promptstyle{plant}{potted plant}}}
\end{tabular}
\caption{\textbf{Failure cases.} Our method is not robust to a few cases such as (a) incomplete or (b) coarse labelling. Another failure mode of our method is when there is an ambiguity in the foreground
class (c-d).}
\vspace{-0.7cm}
\label{fig:failure_cases}
\end{center}
\end{figure}

\subsection{Our method in the wild}
\label{ssec:in_the_wild}

We also test our method in the wild. We randomly download a set of images and provide textual queries we find most suitable. We show in Fig.~\ref{fig:teaser} that our method exhibits off-the-shelf open-world segmentation capabilities, being able to produce masks for specific prompts. More results, including comparisons against other methods of in-the-wild open-world segmentation, are presented in Sec.~\ref{sec:wild_sup} of the Supplementary material.

\section{Conclusions}
We introduce a new method for open-vocabulary semantic segmentation, namely \NAME, which exploits CLIP's open-vocabulary classification abilities.
As opposed to recent approaches we run CLIP densely at multiple scales to obtain coarse semantic mask proposals. When further guided by the quick fully unsupervised object localization method FOUND, which estimates foreground saliency, our model obtains state-of-the-art results on PASCAL VOC and performs on par with baselines on COCO dataset.
Since our method does not require any specific training, it can be used as an off-the-shelf method for open-world segmentation, and could therefore serve as a tool to help dataset annotators. While \NAME already yields competitive results, we believe future work could make it more diverse and efficient.

\section*{Acknowledgments}
We would like to thank Georgy Ponimatkin for the interesting discussions.
This work was supported by the National Centre of Science (Poland) Grant No.2022/45/B/ST6/02817 and by the grant from NVIDIA providing one RTX A5000 24GB used for this project.

{\small
\bibliographystyle{ieee_fullname}
\bibliography{egbib}
}
\clearpage
\newpage
\appendix
\section{Supplementary material}
In this Supplementary material we consider broader impact in Sec.~\ref{sec:broader_impact}, give more details on different foreground-background segmenters used in this work and their adaptations in Sec.~\ref{sec:foreground}. We then show more qualitative results in Sec.~\ref{sec:qualitative_sup}, including comparisons with other methods on the datasets used in the evaluation as well as the examples in the wild. We conclude with an in-depth analysis of the failure case of our method.

\subsection{Broader impact}
\label{sec:broader_impact}
Semantic segmentation plays a crucial role across a wide range of fields, including healthcare, medicine, self-driving cars, and many more. While this technology can foster many applications with a positive impact, there still exists a risk of negative misuse. Additionally, since \NAME builds on foundation models that were trained on large-scale data, our method is not free from biases present in the datasets. Overall \NAME has a broad range of applications. Being training-free, \NAME can especially serve as an off-the-shelf image annotator in computing or budget-limited environments.

\subsection{Foreground-background segmenters}
\label{sec:foreground}
In this section, we provide more details on the foreground-background segmenters we use in our work.
We consider the two variants of FOUND~~\cite{simeoni2023found}: we first use the coarse saliency maps produced
without self-training, noted FOUND-bkg in the paper, which corresponds to the set of similar pixels to the least salient \emph{background seed} pixel in the self-supervised feature space. 
We also use the quick conv1x1 model, named FOUND in the main paper, which is self-trained on only 10,553 images \cite{wang2017learning} and produces more pixel-aligned results. We refer the reader to the original paper for more details.

We also experiment with the state-of-the-art unsupervised panoptic segmentation method CutLER~\cite{wang2023cut}. 
It produces a single mask per discovered object in a scene. We test CutLER in two different setups. 
First, since our method was designed to take one saliency map for the whole image we adapt the output of CutLER to obtain a saliency map as follows. We run CutLER per image obtaining the binary instance masks $\{\zeta_n \in [0, 1]^{H\times W}\}_{n=1..N}$, with $N$ the total number of output binary masks. We also extract the confidence scores corresponding to the masks $\{\sigma_n \in\ \mathbb{R}\}_{n=1..N}$. Note that the output masks are of the size of the input image. We then filter the masks and discard those with a confidence score $\sigma_n < 0.3$ similar to the value on the official CutLER repository \footnote{\url{https://github.com/facebookresearch/CutLER/}}. We then aggregate the remaining masks into a saliency map $ M_{CUT}$ with:

\begin{equation}
\begin{aligned}
    M_{CUT} &= \dfrac{1}{\mathcal{Z}}\sum\limits_{n\in N} \sigma_n\zeta_n, \quad \text{where} \\
    \mathcal{Z} &= \max\limits_{\text{px}} \left(\sum\limits_{n\in N}\sigma_n \zeta_n \right) \in \mathbb{R},
\end{aligned}
\end{equation}

such that $M_{CUT} \in \mathbb{R}^{H\times W}$ is a normalized 2D-mask with its maximum value being $1$. 

Second, for a fair comparison, we also use CutLER off-the-shelf as a mask extractor. We use previously described masks $\zeta_n \in [0, 1]^{H\times W}$ and with each one of them, we create an image $I_n$ mask where the background is masked out. Each masked image $I_n$ is then fed separately to CLIP to obtain a CLIP prediction. We denote this approach as \emph{CutLER mask} in Tab.~\ref{tab:saliency} of the main paper. 

Overall, \NAME achieves the best performances with the light self-trained FOUND as discussed in the main paper.

\subsection{More qualitative results}
\label{sec:qualitative_sup}
We provide in this section more qualitative examples produced with \NAME. We first compare our method against other state-of-the-art approaches in Fig.~\ref{fig:results_sup_pascal} for PASCAL VOC and Fig.~\ref{fig:results_sup_coco} for COCO.  

In Fig.~\ref{fig:in_the_wild_supp}, we then present more in-the-wild examples and conclude this section by discussing failure cases and limitations of our method in detail. 

\subsubsection{Comparisons}
\setlength{\pascalwidth}{0.16\textwidth}
\setlength{\resultsheight}{2.03cm}

\begin{figure*}[h]
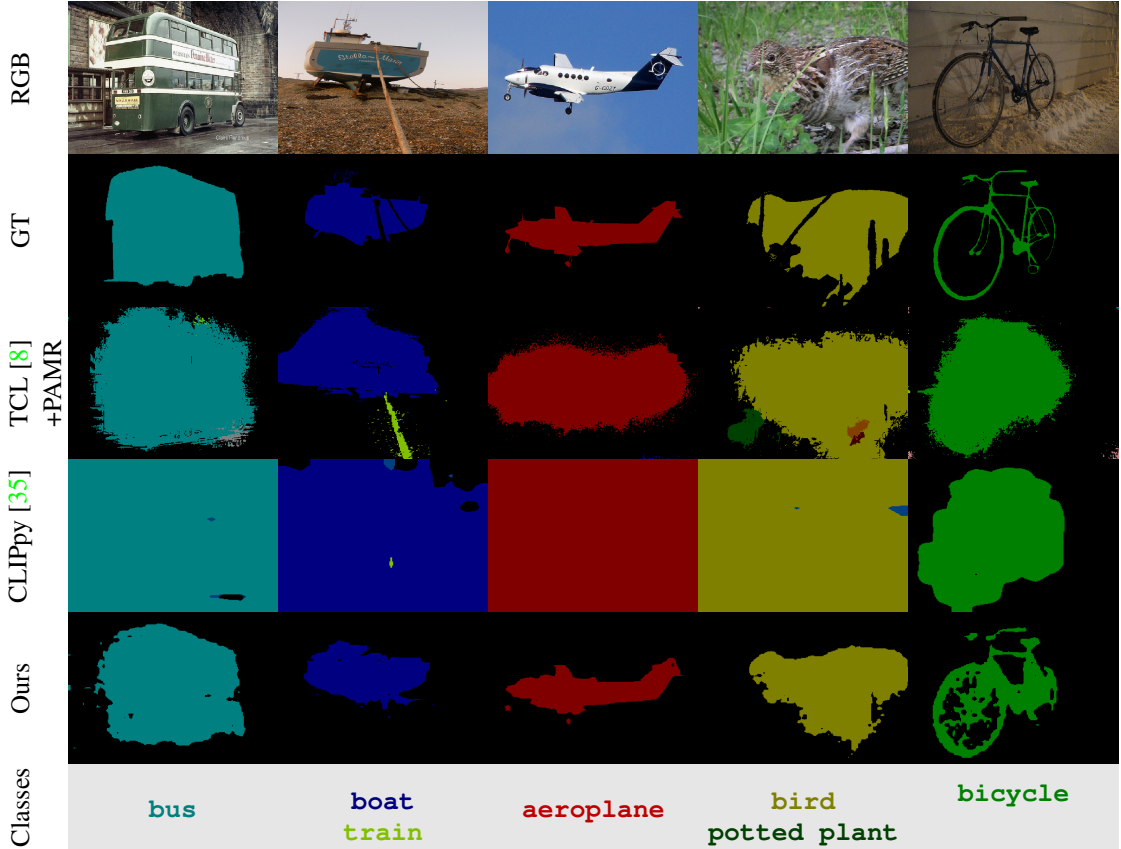

\centering
\renewcommand{\arraystretch}{0}
\setlength\tabcolsep{0pt} 

\begin{center}
\begin{tabular}{
>{\centering\arraybackslash}m{1.2em} 
>{\centering\arraybackslash}m{1.2em} 
>{\centering\arraybackslash}m{\pascalwidth}
>{\centering\arraybackslash}m{\pascalwidth}
>{\centering\arraybackslash}m{\pascalwidth}
>{\centering\arraybackslash}m{\pascalwidth}
>{\centering\arraybackslash}m{\pascalwidth}
}

\rotatebox{90}{RGB} & & 
\pascalresult{pascal_2011_001341}{img} &
\pascalresult{pascal_2009_000446}{img} &
\pascalresult{pascal_2009_001851}{img} &
\pascalresult{pascal_2010_003362}{img} & 
\pascalresult{pascal_2010_000342}{img} \\

\rotatebox{90}{GT} & &
\pascalresult{pascal_2011_001341}{gt} &
\pascalresult{pascal_2009_000446}{gt} &
\pascalresult{pascal_2009_001851}{gt} &
\pascalresult{pascal_2010_003362}{gt} &
\pascalresult{pascal_2010_000342}{gt} \\


\rotatebox{90}{TCL~\cite{cha2022tcl}}  & \rotatebox{90}{+PAMR} & 
\pascalresult{pascal_2011_001341}{tcl} &
\pascalresult{pascal_2009_000446}{tcl} &
\pascalresult{pascal_2009_001851}{tcl} &
\pascalresult{pascal_2010_003362}{tcl} & 
\pascalresult{pascal_2010_000342}{tcl} \\

\rotatebox{90}{CLIPpy~\cite{clippy2022}} & &
\pascalresult{pascal_2011_001341}{clippy} &
\pascalresult{pascal_2009_000446}{clippy} &
\pascalresult{pascal_2009_001851}{clippy} &
\pascalresult{pascal_2010_003362}{clippy} & 
\pascalresult{pascal_2010_000342}{clippy} \\


\rotatebox{90}{Ours} & &
\pascalresult{pascal_2011_001341}{pred} &
\pascalresult{pascal_2009_000446}{pred} &
\pascalresult{pascal_2009_001851}{pred} &
\pascalresult{pascal_2010_003362}{pred} &
\pascalresult{pascal_2010_000342}{pred} \\

\vspace{1pt}
 \rotatebox{90}{Classes} & ~ & 
 {\cellcolor[gray]{.9} \makecell{\promptstyle{bus}{bus} 
}}
&
{\cellcolor[gray]{.9}\makecell{\promptstyle{boat}{boat} \\ \promptstyle{train}{train}}}
&
 {\cellcolor[gray]{.9}\makecell
{\promptstyle{aeroplane}{aeroplane}}}
&
{\cellcolor[gray]{.9}\makecell 
{\promptstyle{bird_pascal}{bird} \\ \promptstyle{plant_pascal}{potted plant}}}
& 
{\cellcolor[gray]{.9}\makecell{\promptstyle{bike}{bicycle}}}

\end{tabular}
\end{center}
    \vspace{-1.5em}
    \caption{
    \textbf{Qualitative segmentation results on PASCAL VOC}. We compare our method against CLIPpy~\cite{clippy2022} and TCL (with PAMR post-processing)~\cite{cha2022tcl}. Our method consistently outperforms two other methods by producing accurate segmentation masks.}
    \label{fig:results_sup_pascal}
\end{figure*}
We first present comparisons with other methods on the two segmentation datasets used in this work, namely PASCAL VOC~\cite{pascal-voc-2012} and COCO Object~\cite{lin2014microsoft}.

\paragraph{PASCAL VOC} 
Fig.~\ref{fig:results_sup_pascal} shows randomly sampled images from PASCAL VOC dataset and the results of \NAME and our baselines.
Our method produces accurate masks for all of the images and the result of \NAME is the closest to the ground truth compared to other methods. We observe that the two other methods, TCL~\cite{cha2022tcl} and CLIPpy~\cite{clippy2022}, produce masks that are too coarse, with the latter frequently even assigning most of the image to one segment.

\newlength{\resultscocoheight}
\setlength{\resultscocoheight}{2.0cm}

\newcommand{\cocoresultt}[2]{%
\begin{minipage}[b]{\cocowidth}
     \centering
     \includegraphics[width=\textwidth, height=\resultscocoheight]{figures/images/result_samples/#1_#2.png}
\end{minipage}}
\newcommand{\cocoresulttjpeg}[2]{%
\begin{minipage}[b]{\cocowidth}
     \centering
     \includegraphics[width=\textwidth, height=\resultscocoheight]{figures/images/result_samples/#1_#2.jpg}
\end{minipage}}

\begin{figure*}[h]
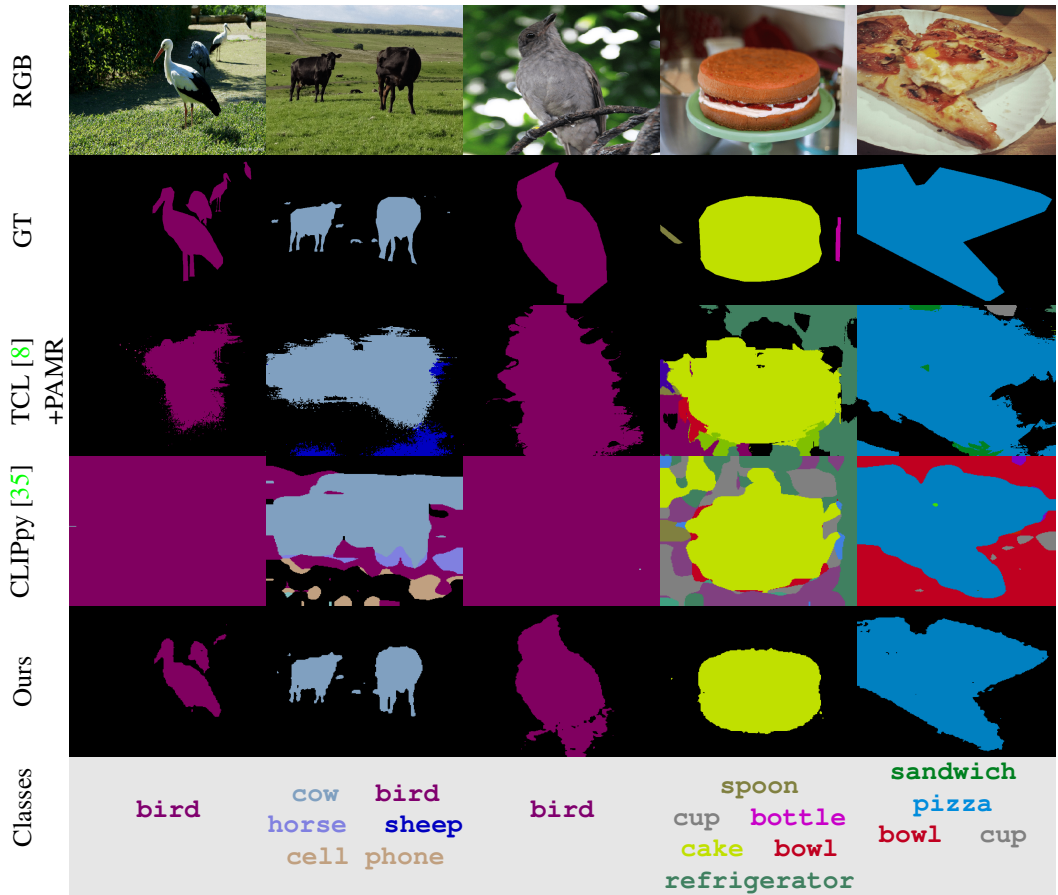

\centering
\renewcommand{\arraystretch}{0}
\setlength\tabcolsep{0pt} 

\begin{center}
\begin{tabular}{
>{\centering\arraybackslash}m{1.2em} 
>{\centering\arraybackslash}m{1.2em} 
>{\centering\arraybackslash}m{\cocowidth}
>{\centering\arraybackslash}m{\cocowidth}
>{\centering\arraybackslash}m{\cocowidth}
>{\centering\arraybackslash}m{\cocowidth}
>{\centering\arraybackslash}m{\cocowidth}
}

\rotatebox{90}{RGB} & & 
\cocoresultt{coco_505565}{img} &
\cocoresultt{coco_244411}{img} &
\cocoresultt{coco_56545}{img} &
\cocoresultt{coco_89670}{img} & 
\cocoresultt{coco_232489}{img} \\

\rotatebox{90}{GT} & &
\cocoresultt{coco_505565}{gt} &
\cocoresultt{coco_244411}{gt} &
\cocoresultt{coco_56545}{gt} &
\cocoresultt{coco_89670}{gt} & 
\cocoresultt{coco_232489}{gt} \\


\rotatebox{90}{TCL~\cite{cha2022tcl}}  & \rotatebox{90}{+PAMR} & 
\cocoresultt{coco_505565}{tcl} &
\cocoresultt{coco_244411}{tcl} &
\cocoresultt{coco_56545}{tcl} &
\cocoresultt{coco_89670}{tcl} & 
\cocoresultt{coco_232489}{tcl} \\

\rotatebox{90}{CLIPpy~\cite{clippy2022}} & &
\cocoresultt{coco_505565}{clippy} &
\cocoresultt{coco_244411}{clippy} &
\cocoresultt{coco_56545}{clippy} &
\cocoresultt{coco_89670}{clippy} & 
\cocoresultt{coco_232489}{clippy} \\


\rotatebox{90}{Ours} & &
\cocoresultt{coco_505565}{pred} &
\cocoresultt{coco_244411}{pred} &
\cocoresultt{coco_56545}{pred} &
\cocoresultt{coco_89670}{pred} & 
\cocoresultt{coco_232489}{pred} \\

\vspace{1pt}
 \rotatebox{90}{Classes} & ~ & 
 {\cellcolor[gray]{.9} \makecell{ 
 \promptstyle{bird_coco}{bird}
}}
&
{\cellcolor[gray]{.9} \makecell {\promptstyle{cow_coco}{cow} $\>\>\>\>$ \promptstyle{bird_coco}{bird} \\ \promptstyle{horse}{horse} $\>\>\>\>$ \promptstyle{sheep_coco}{sheep} \\ \promptstyle{cellphone}{cell phone}}}
&
 {\cellcolor[gray]{.9}\makecell
{\promptstyle{bird_coco}{bird}}}
&
{\cellcolor[gray]{.9}\makecell 
{\promptstyle{spoon}{spoon} \\ \promptstyle{cup}{cup} $\>\>\>$ \promptstyle{bottle}{bottle} \\\promptstyle{cake}{cake} $\>\>\>$ \promptstyle{bowl}{bowl}
\\
\promptstyle{refrigerator}{refrigerator}}}
& 
{\cellcolor[gray]{.9}
\makecell{\promptstyle{sandwich}{sandwich} \\  \promptstyle{pizza}{pizza} \\ 
\promptstyle{bowl}{bowl} $\>\>\>\>$ \promptstyle{cup}{cup}}
}

\end{tabular}
\end{center}
    \vspace{-1.5em}
    \caption{
    \textbf{Qualitative segmentation results on COCO}. We compare our method against CLIPpy~\cite{clippy2022} and TCL~\cite{cha2022tcl} (with PAMR~\cite{pamr} post-processing)~\cite{cha2022tcl}. Our method consistently outperforms two other methods by producing accurate segmentation masks. TCL and CLIPpy also both suffer from hallucinating or producing noisy masks.
    }
    \label{fig:results_sup_coco}
\end{figure*}
\paragraph{COCO Object} Fig.~\ref{fig:results_sup_coco} shows the examples from COCO dataset. 
While generating masks with mostly the correct category, TCL produces very noisy boundaries compared to \NAME. CLIPpy not only generates noisy masks but also produces a lot of clutter, assigning often wrong labels to background pixels.

\subsubsection{In-the-wild examples}
\label{sec:wild_sup}
\begin{figure*}[!t]
\centering
\renewcommand{\arraystretch}{0} 
\setlength\tabcolsep{0pt}  
\begin{center}
\begin{tabular}{
>{\centering\arraybackslash}m{1em}
>{\centering\arraybackslash}m{0.22\textwidth}
>{\centering\arraybackslash}m{0.23\textwidth}
>{\centering\arraybackslash}m{0.32\textwidth}
}
\rotatebox{90}{RGB} &
\includegraphics[width=\linewidth, height=2.5cm]{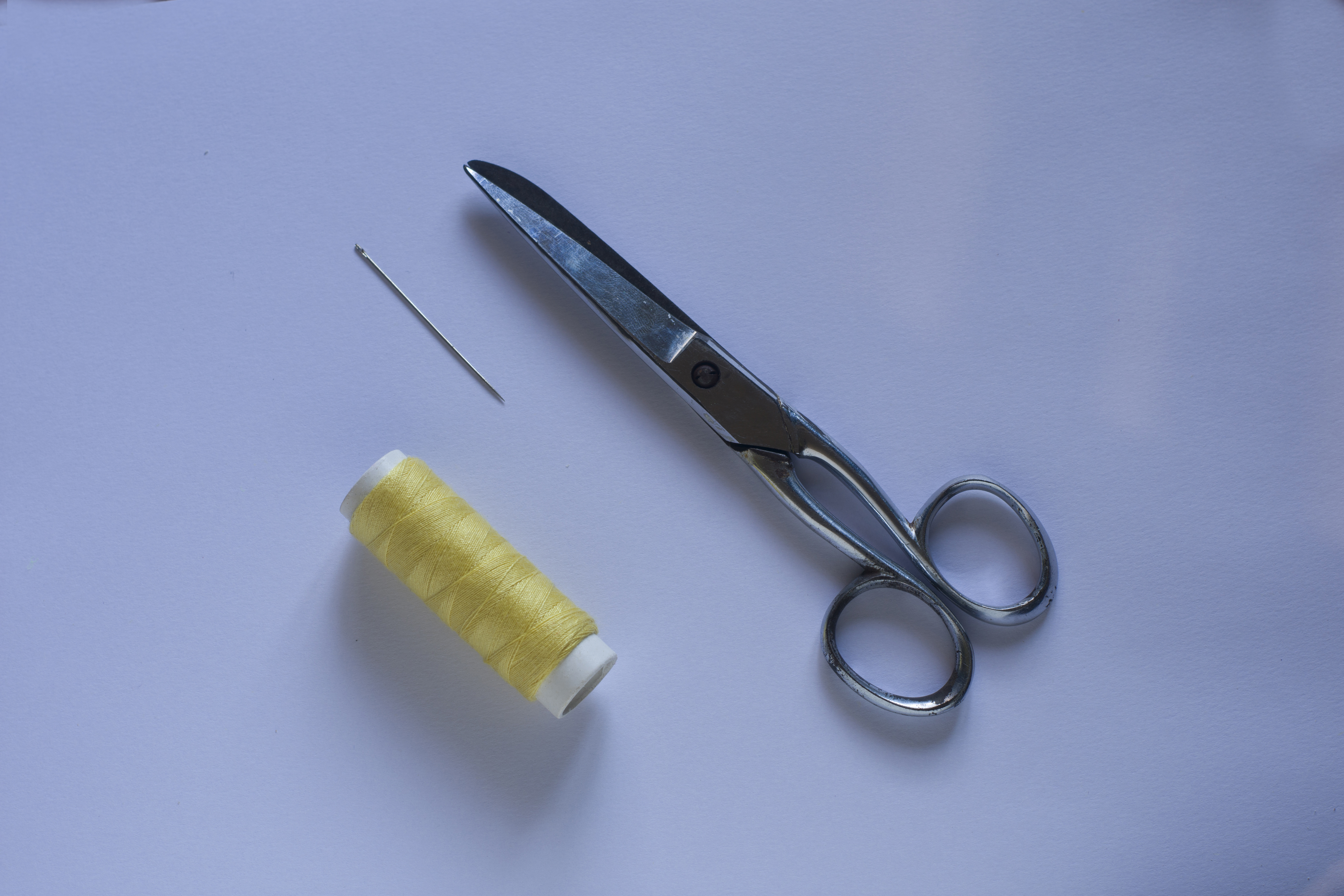} 
& 
\includegraphics[width=\linewidth, height=2.5cm]{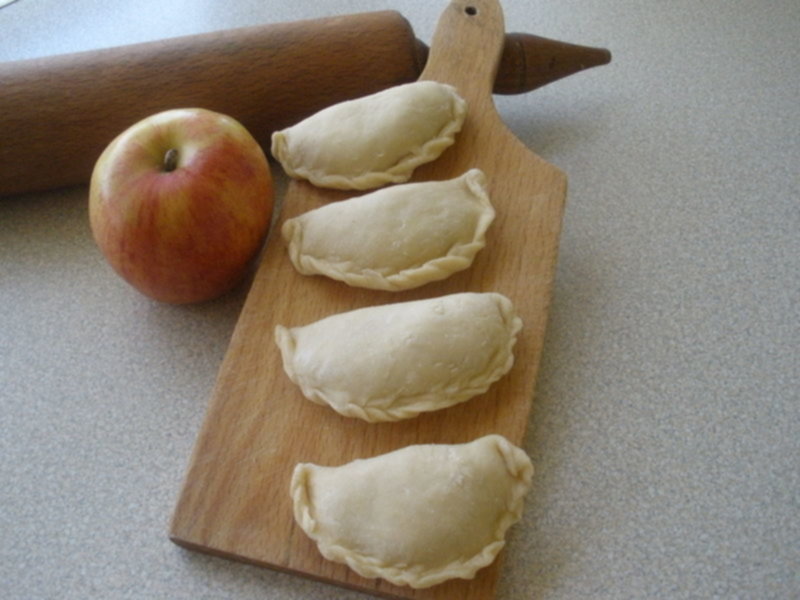} 
& 
\includegraphics[width=\linewidth, height=2.5cm,trim=10 60 10 30, clip]{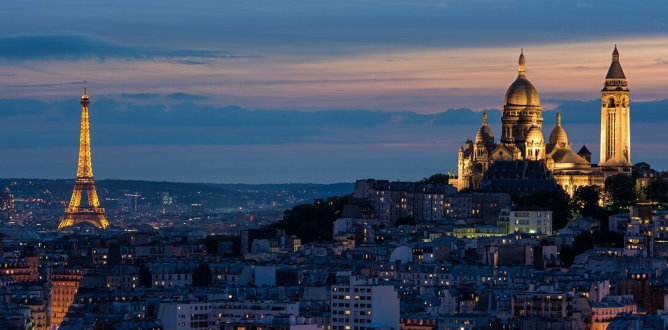} \\

\rotatebox{90}{TCL~\cite{cha2022tcl} +PAMR} &
\includegraphics[width=\linewidth, height=2.5cm]{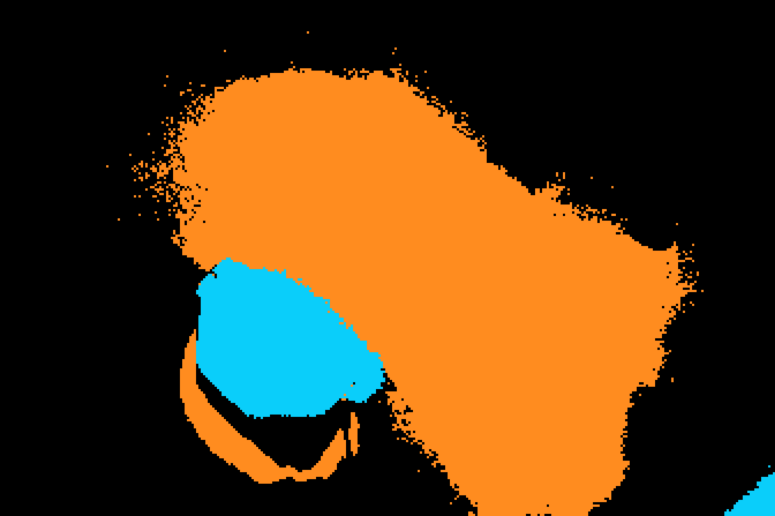}
& 
\includegraphics[width=\linewidth, height=2.5cm]{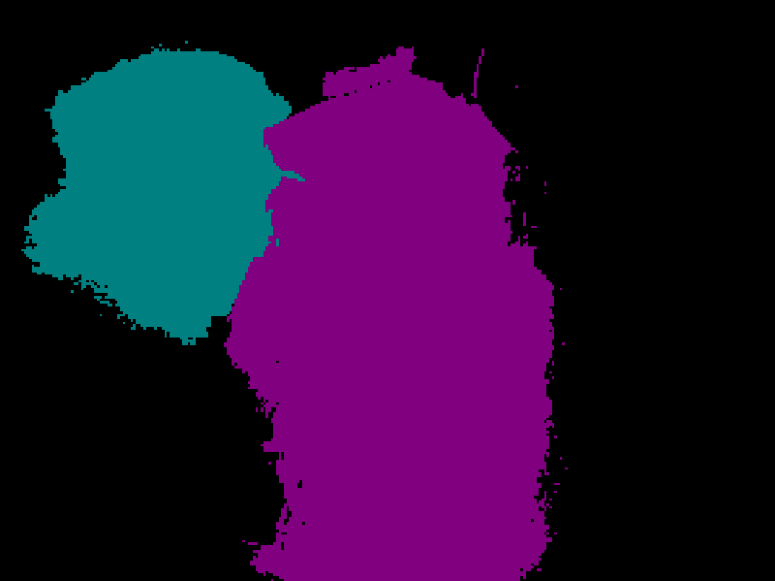} 
& 
\includegraphics[width=\linewidth, height=2.5cm,trim=10 60 10 30, clip]{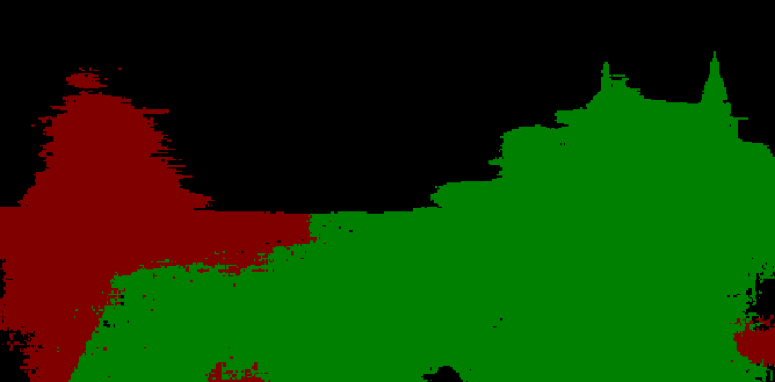} \\

\rotatebox{90}{\NAME} &
\includegraphics[width=\linewidth, height=2.5cm]{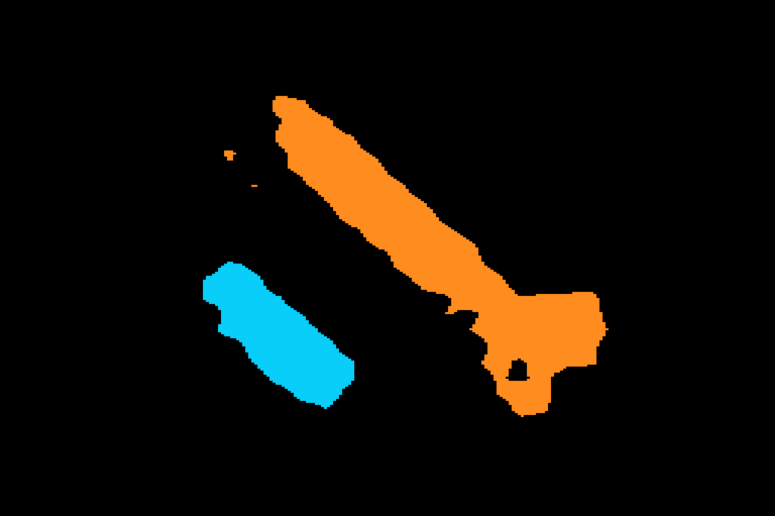} 
& 
\includegraphics[width=\linewidth,  height=2.5cm]{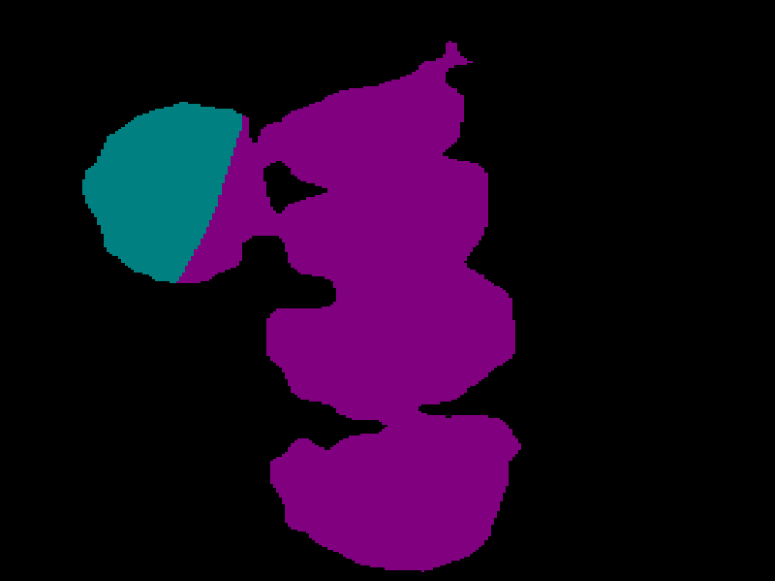} 
& 
\includegraphics[width=\linewidth,  height=2.5cm,trim=10 60 10 30, clip]{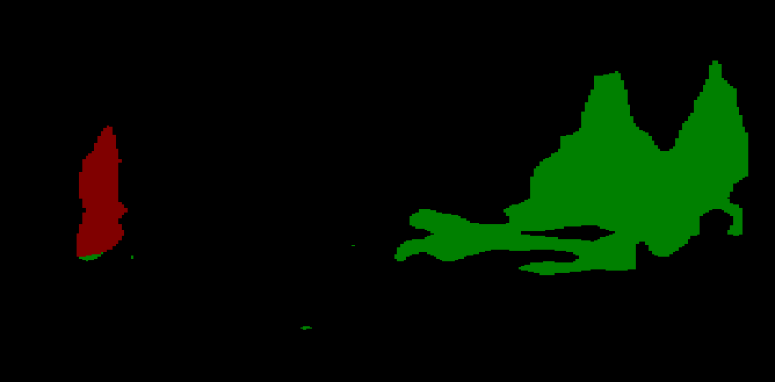} \\
~ &
{\cellcolor[gray]{.9} \makecell{
\promptstyle{scissors}{metal scissors} \\ \promptstyle{thread}{thread} }
}
& 
{\cellcolor[gray]{.9} \makecell{\promptstyle{apple}{apple} \\
\promptstyle{dumplings}{polish dumplings}}
}
&
{\cellcolor[gray]{.9} \makecell{\promptstyle{red}{Eiffel tower} $\>\>\>\>$ \promptstyle{green}{Sacré coeur}} }\\

\rotatebox{90}{RGB} &
\includegraphics[width=\linewidth,  height=2.8cm]{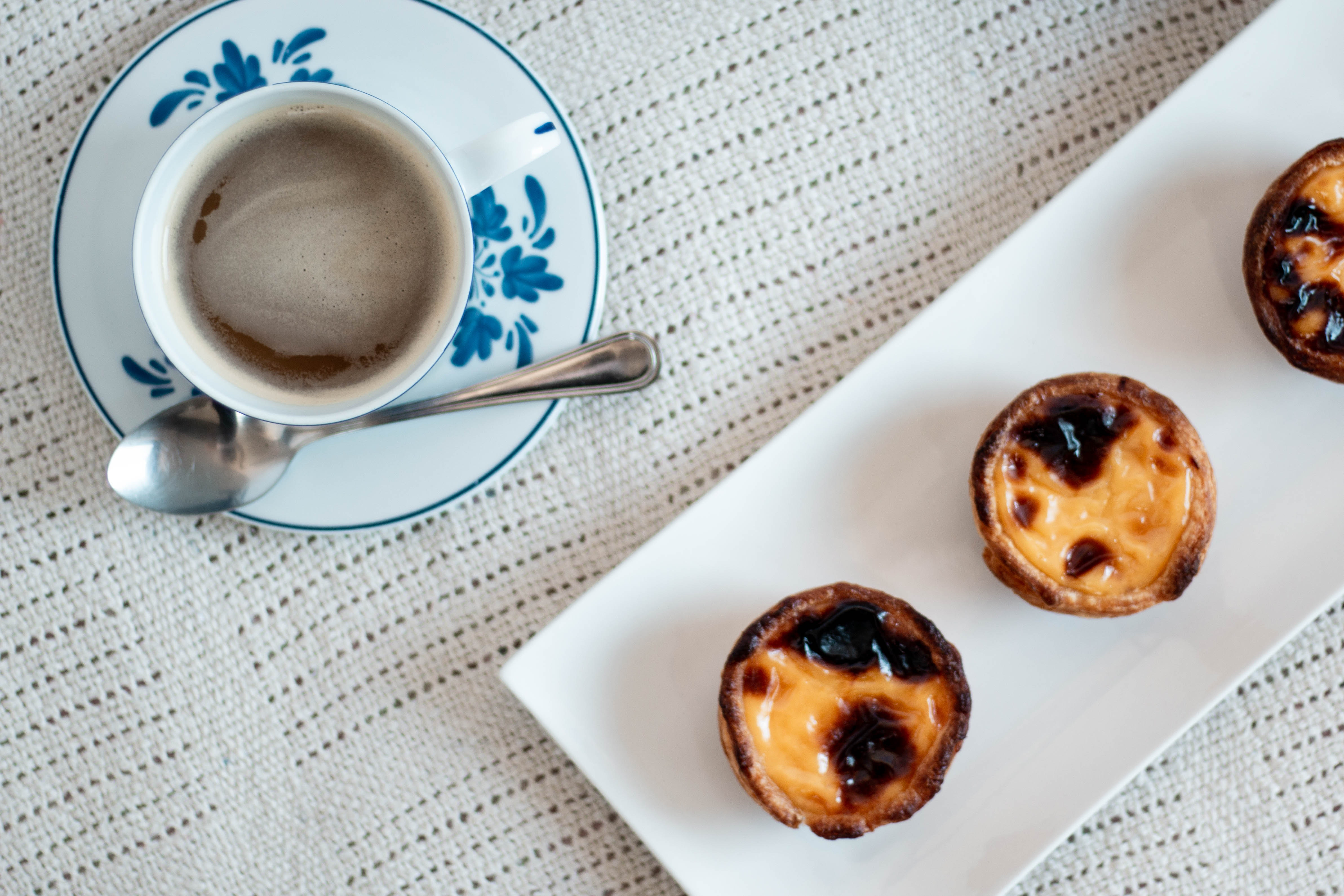} 
& 
\includegraphics[width=\linewidth,  height=2.8cm]{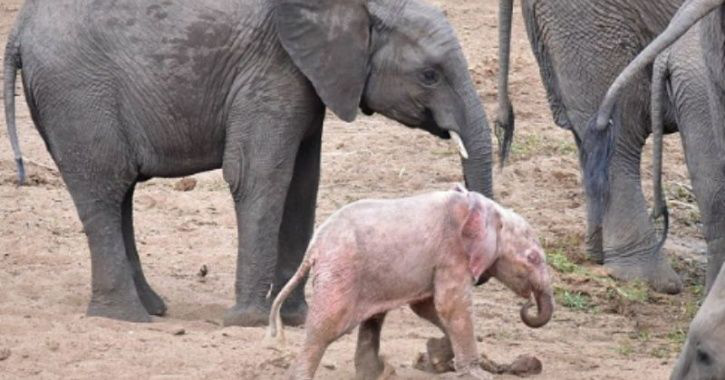} 
& 
\includegraphics[width=\linewidth,  height=2.8cm]{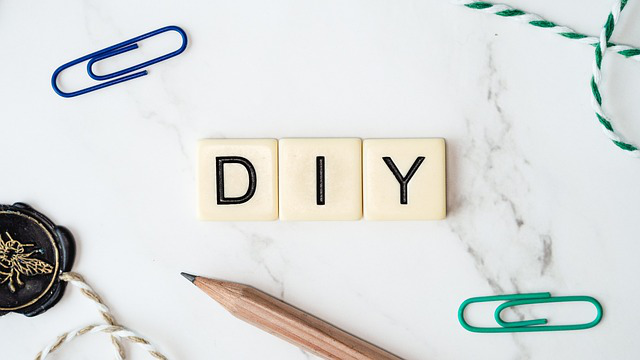} \\

\rotatebox{90}{TCL~\cite{cha2022tcl}+PAMR~\cite{pamr}} &
\includegraphics[width=\linewidth,  height=2.8cm]{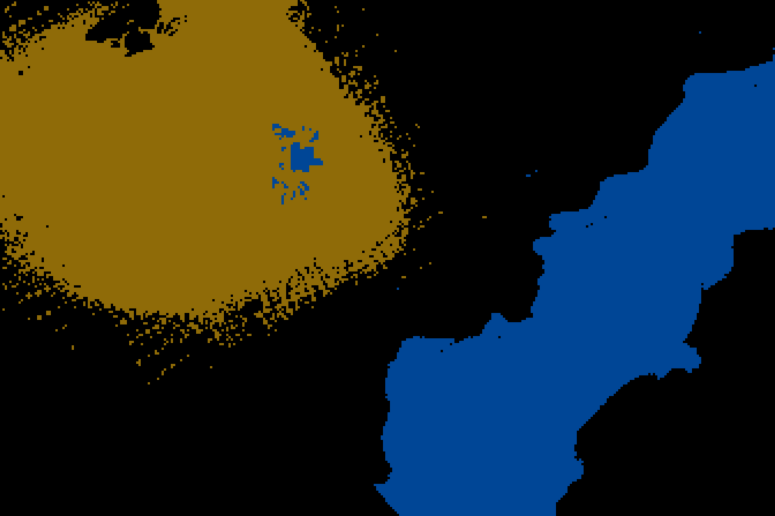}
& 
\includegraphics[width=\linewidth,  height=2.8cm]{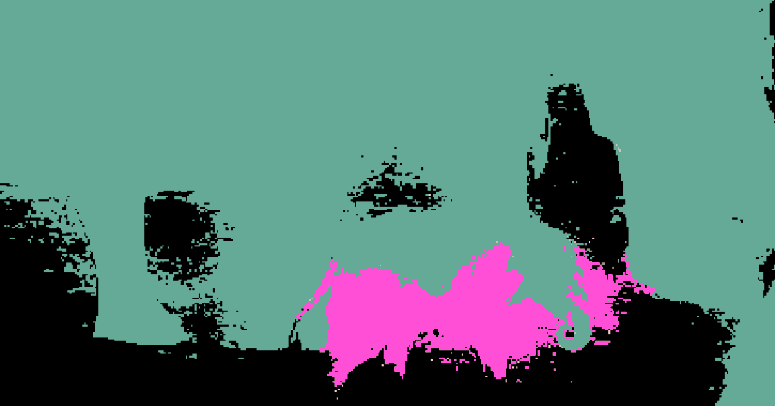} 
& 
\includegraphics[width=\linewidth,  height=2.8cm]{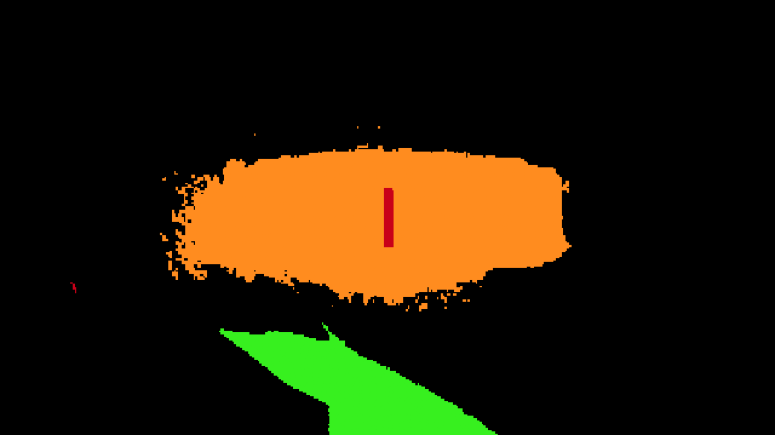} \\

\rotatebox{90}{\NAME} &
\includegraphics[width=\linewidth,  height=2.8cm]{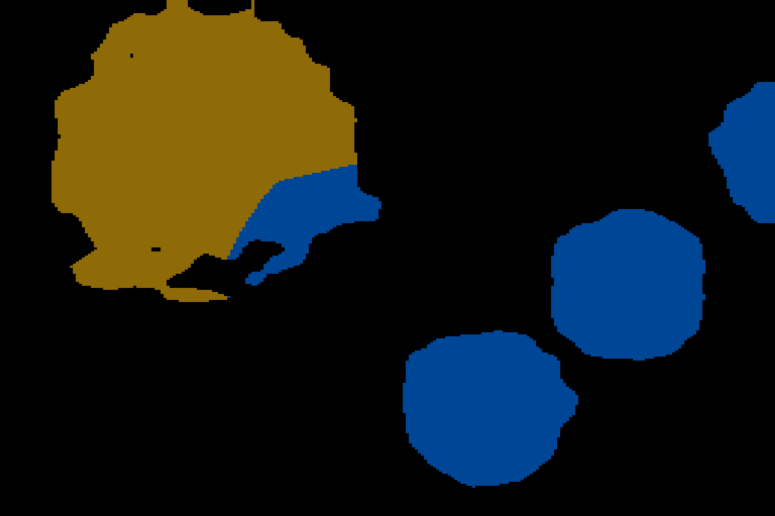} 
& 
\includegraphics[width=\linewidth,  height=2.8cm]{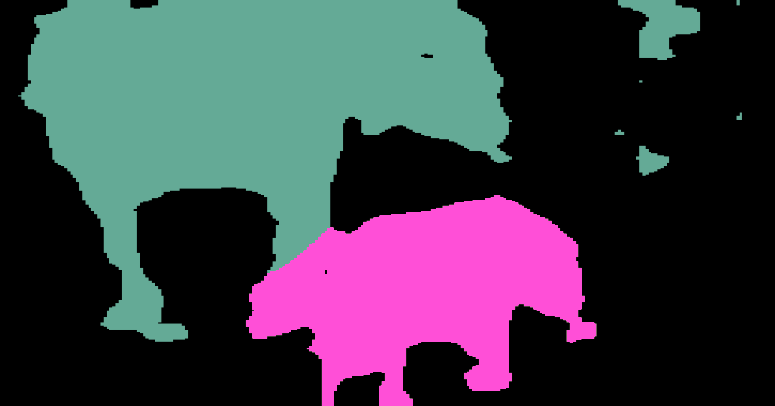} 
& 
\includegraphics[width=\linewidth,  height=2.8cm]{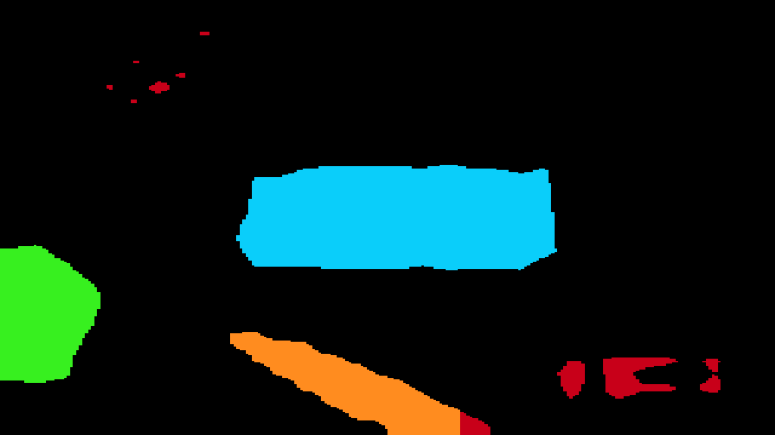} \\
~ &
{\cellcolor[gray]{.9} \makecell{
\promptstyle{pasteis}{pasteis de nata} \\ \promptstyle{coffee}{coffee}}
}
& 
{\cellcolor[gray]{.9} \makecell{\promptstyle{greyeleph}{grey elephant} \\
\promptstyle{pinkeleph}{pink elephant}}
}
&
{\cellcolor[gray]{.9} \makecell{\promptstyle{scrabble}{scrabble} $\>\>\>\>$ \promptstyle{paper clip}{paper clip} \\
\promptstyle{rubber stamp}{rubber stamp} $\>\>\>\>$ \promptstyle{pencil}{pencil}}}

\end{tabular}
    \caption{\textbf{More examples of in-the-wild open-world segmentation.} We compare segmentation produced by our method with the results of TCL~\cite{cha2022tcl}. While both methods are able to detect and locate each class, including distinguishing between \promptstyle{pinkeleph}{pink elephant} and \promptstyle{greyeleph}{grey elephant}, TCL largely over-segments objects.
    }
    \label{fig:in_the_wild_supp}
\end{center}
\end{figure*}

We provide more in-the-wild examples to showcase the open-vocabulary abilities of our method. In Fig.~\ref{fig:in_the_wild_supp} we present a couple of randomly mined images from the Web in comparison with TCL~\cite{cha2022tcl}. Both of the methods correctly assign queries to proper segments, even very specific types of objects, such as traditional dishes e.g. \promptstyle{dumplings}{polish dumplings} and \promptstyle{pasteis}{pasteis de nata}; monuments \promptstyle{red}{Eiffel tower} and \promptstyle{green}{Sacré coeur}. Moreover, thanks to CLIP backbone both methods can distinguish between different colours, e.g. \promptstyle{greyeleph}{grey elephant} against \promptstyle{pinkeleph}{pink elephant}. However, we observe that the quality of masks produced by TCL again is not as detailed as ours. Note that TCL uses PAMR post-processing technique thus we would expect the generated masks to be more precise. 

\subsubsection{Failure cases}
\label{sec:failures_sup}
We analyze failure cases of our method in Fig.~\ref{fig:failure_sup_cases}. We can see that \NAME suffers from producing incomplete masks column (a) and missing objects (c). This happens due to the saliency produced by the foreground-background segmenter, which in the case of complex, multi-object scenes focuses on certain aspects of a scene. Moreover, our method has limited performance in the case of overlapping objects, such as \promptstyle{dog}{dog} and \promptstyle{chair_coco}{chair} in column (d).
Finally, we find more failures due to inaccurate annotations, such as the one in column (b), where \promptstyle{bowl}{bowl} is misclassified by what is inside, i.e. \promptstyle{carrot}{carrot}.
\newlength{\failureheight}
\setlength{\failureheight}{2.6cm}
\newcommand{\pascalresulttwo}[2]{%
\begin{minipage}[b]{0.99\linewidth}
     \centering
     \includegraphics[width=\textwidth,height=\failureheight]{figures/images/result_samples/pascal_voc/#1_#2.png}
\end{minipage}}
\newcommand{\pascalresulttwojpeg}[2]{%
\begin{minipage}[b]{0.99\linewidth}
     \centering
     \includegraphics[width=\textwidth,height=\failureheight]{figures/images/result_samples/pascal_voc/#1_#2.jpg}
\end{minipage}}

\begin{figure*}[htbp]
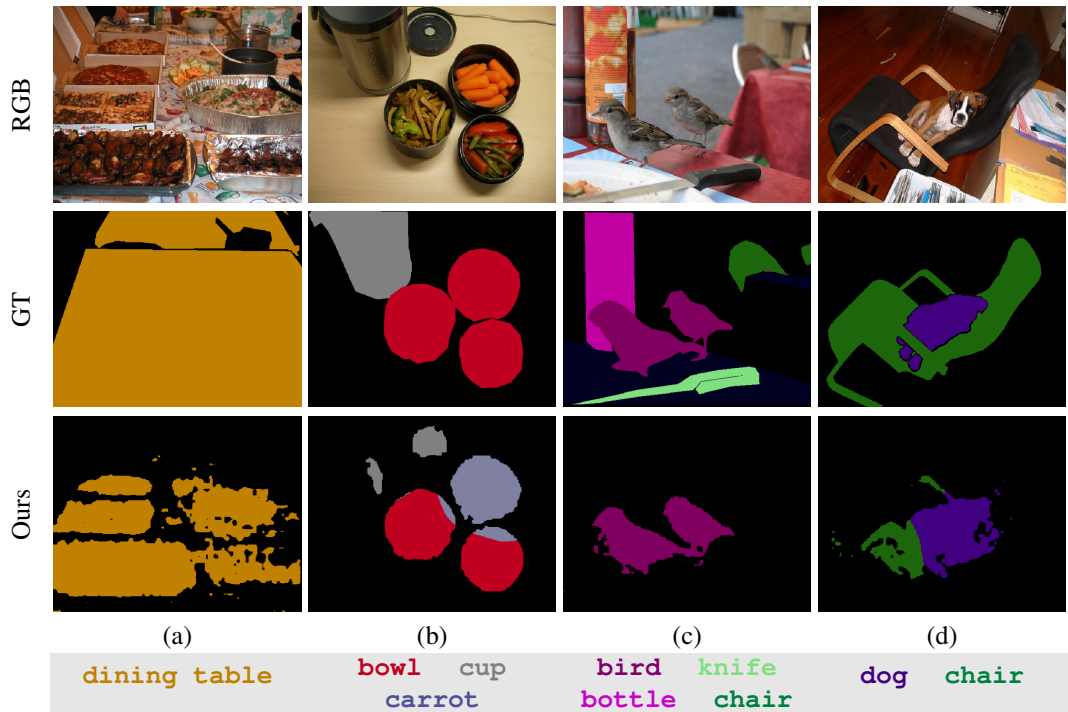

\centering
\renewcommand{\arraystretch}{1} 
\setlength\tabcolsep{1pt}  
\begin{center}
\begin{tabular}{
>{\centering\arraybackslash}m{2em}
>{\centering\arraybackslash}m{0.19\linewidth}
>{\centering\arraybackslash}m{0.19\linewidth}
>{\centering\arraybackslash}m{0.19\linewidth}
>{\centering\arraybackslash}m{0.19\linewidth}}

\rotatebox{90}{RGB} &
\pascalresulttwo{pascal_2008_008051}{img} &
\cocoresulttwo{coco_194940}{img} & 
\cocoresulttwo{coco_277051}{img} &
\pascalresulttwo{pascal_2010_000003}{img} 
\\
\rotatebox{90}{GT} &
\pascalresulttwo{pascal_2008_008051}{gt}
& \cocoresulttwo{coco_194940}{gt} 
& \cocoresulttwo{coco_277051}{gt}
&\pascalresulttwo{pascal_2010_000003}{gt}  \\
\rotatebox{90}{Ours} & 
\pascalresulttwo{pascal_2008_008051}{pred} &
\cocoresulttwo{coco_194940}{pred} 
&\cocoresulttwo{coco_277051}{pred}
& \pascalresulttwo{pascal_2010_000003}{pred} \\
~ & (a) & (b) & (c) & (d) \\
~ 
& 
{\cellcolor[gray]{.9} \makecell{\promptstyle{diningtable}{dining table} }
}
&
{\cellcolor[gray]{.9} \makecell{\promptstyle{bowl}{bowl} $\>\>\>\>$ \promptstyle{cup}{cup} \\
\promptstyle{carrot}{carrot}} }
& 
{\cellcolor[gray]{.9} \makecell{
\promptstyle{bird_coco}{bird} $\>\>\>\>$ \promptstyle{knife}{knife}\\ \promptstyle{bottle}{bottle}
 $\>\>\>\>$ \promptstyle{chair_coco}{chair}}
}
&
{\cellcolor[gray]{.9} \makecell{\promptstyle{dog}{dog} $\>\>\>\>$ \promptstyle{chair_coco}{chair}} }
 \\
\end{tabular}
\caption{\textbf{Failure cases.} We show examples from both datasets. Our method at times produces incomplete masks when saliency focuses only on parts of the scene such as in (a) and (c), ambiguous classification in (b), as well confusion when classes overlap (d).  }
\label{fig:failure_sup_cases}
\end{center}
\end{figure*}

\subsubsection{Detailed quantitative results}
\label{sec:class_sup}
\begin{table*}[tbp]
\centering
\caption{\NAME zero-shot performance (IoU) on the 21 classes from Pascal VOC. The background class is denoted as \faBackground.}
\label{tab:classes}
\begin{adjustbox}{width=0.98\textwidth}
  \begin{tabular}{r@{\hskip1em}*{21}{c}@{\hskip1em}}
  \toprule
   & \tableIcons \\ \midrule	
    & 87.6& 78.1& 33.8& 77.5& 62.6& 65.4& 71.4& 66.0& 81.6& 16.1& 75.2& 20.0& 78.5& 69.8& 63.8& 56.6& 37.3& 79.9& 25.0& 68.3& 37.5\\
  \bottomrule
  \end{tabular}
\end{adjustbox}
\vspace{-1em}
\end{table*}

We present detailed quantitative results on PASCAL VOC in Tab.~\ref{tab:classes} for each class. We observe that the worst performance is obtained on classes which are typically only partially visible in images, such as furniture (chair, sofa and table). This is mostly due to the decreased performance of the saliency detector in those classes, which is biased towards object-centric images. The high performance (87.6 IoU) for the \texttt{background} class confirms the efficacy of our saliency detector.

\end{document}